\documentclass[letterpaper]{article} % DO NOT CHANGE THIS
\usepackage{aaai23}
\usepackage{times}  % DO NOT CHANGE THIS
\usepackage{helvet}  % DO NOT CHANGE THIS
\usepackage{courier}  % DO NOT CHANGE THIS
\usepackage[hyphens]{url}  % DO NOT CHANGE THIS
\usepackage{graphicx} % DO NOT CHANGE THIS
\usepackage{natbib}  % DO NOT CHANGE THIS AND DO NOT ADD ANY OPTIONS TO IT
\usepackage{cite}
\usepackage{multirow}
\usepackage[title]{appendix}
\usepackage{algorithm}
\usepackage{algorithmic}
\usepackage{comment}
\usepackage{graphicx}
\usepackage{amsmath}
\usepackage{helvet}  % DO NOT CHANGE THIS
\usepackage{courier}  % DO NOT CHANGE THIS
\usepackage[hyphens]{url}  % DO NOT CHANGE THIS
\usepackage[hidelinks]{hyperref}
\usepackage{graphicx} % DO NOT CHANGE THIS
\usepackage{cite}
\usepackage{natbib}  % DO NOT CHANGE THIS AND DO NOT ADD ANY OPTIONS TO IT
\usepackage{caption} % DO NOT CHANGE THIS AND DO NOT ADD ANY OPTIONS TO IT
\DeclareCaptionStyle{ruled}{labelfont=normalfont,labelsep=colon,strut=off} % DO NOT CHANGE THIS
\usepackage{booktabs}
\usepackage{multirow}
\usepackage{color}
\usepackage{float}
\usepackage{graphicx}
\usepackage{cuted}
\newcommand{\red}[1]{{\leavevmode\color{red}[#1]}}

\usepackage{amssymb}
\usepackage[normalem]{ulem}
\usepackage[normalem]{ulem}
\usepackage{multirow}
\usepackage{colortbl}
\usepackage{caption} % DO NOT CHANGE THIS AND DO NOT ADD ANY OPTIONS TO IT
\definecolor{renjiao}{RGB}{0,139,139}

\usepackage{algorithm}
\usepackage{algorithmic}
\usepackage{amsmath}
\usepackage{booktabs}
\usepackage{newfloat}
\usepackage{listings}
\DeclareCaptionStyle{ruled}{labelfont=normalfont,labelsep=colon,strut=off} % DO NOT CHANGE THIS
\floatstyle{ruled}
\newfloat{listing}{tb}{lst}{}
\floatname{listing}{Listing}
\title{Multi-resolution Monocular Depth Map Fusion by Self-supervised \\Gradient-based Composition}
\author{
	Yaqiao Dai\footnote{Co-first authors.}, Renjiao Yi\footnotemark[1], Chenyang Zhu\footnote{Corresponding authors: chenyang.chandler.zhu@gmail.com, kevin.kai.xu@gmail.com. }, Hongjun He, Kai Xu\footnotemark[2]
}
\affiliations{
	National University of Defense Technology}
\usepackage{bibentry}
\begin{document}
	
	\maketitle
	
	\begin{abstract}
		Monocular depth estimation is a challenging problem on which deep neural networks have demonstrated great potential. However, depth maps predicted by existing deep models usually lack fine-grained details due to the convolution operations and the down-samplings in networks. We find that increasing input resolution is helpful to preserve more local details while the estimation at low resolution is more accurate globally. Therefore, we propose a novel depth map fusion module to combine the advantages of estimations with multi-resolution inputs. Instead of merging the low- and high-resolution estimations equally, we adopt the core idea of Poisson fusion, trying to implant the gradient domain of high-resolution depth into the low-resolution depth. While classic Poisson fusion requires a fusion mask as supervision, we propose a self-supervised framework based on guided image filtering. We demonstrate that this gradient-based composition performs much better at noisy immunity, compared with the state-of-the-art depth map fusion method. Our lightweight depth fusion is one-shot and runs in real-time, making our method 80X faster than a state-of-the-art depth fusion method. Quantitative evaluations demonstrate that the proposed method can be integrated into many fully convolutional monocular depth estimation backbones with a significant performance boost, leading to state-of-the-art results of detail enhancement on depth maps. Codes are released at {\color{green}{\url{https://github.com/yuinsky/gradient-based-depth-map-fusion}}}. 
	\end{abstract}

	\begin{figure}
		\centering
		\includegraphics[width=\linewidth]{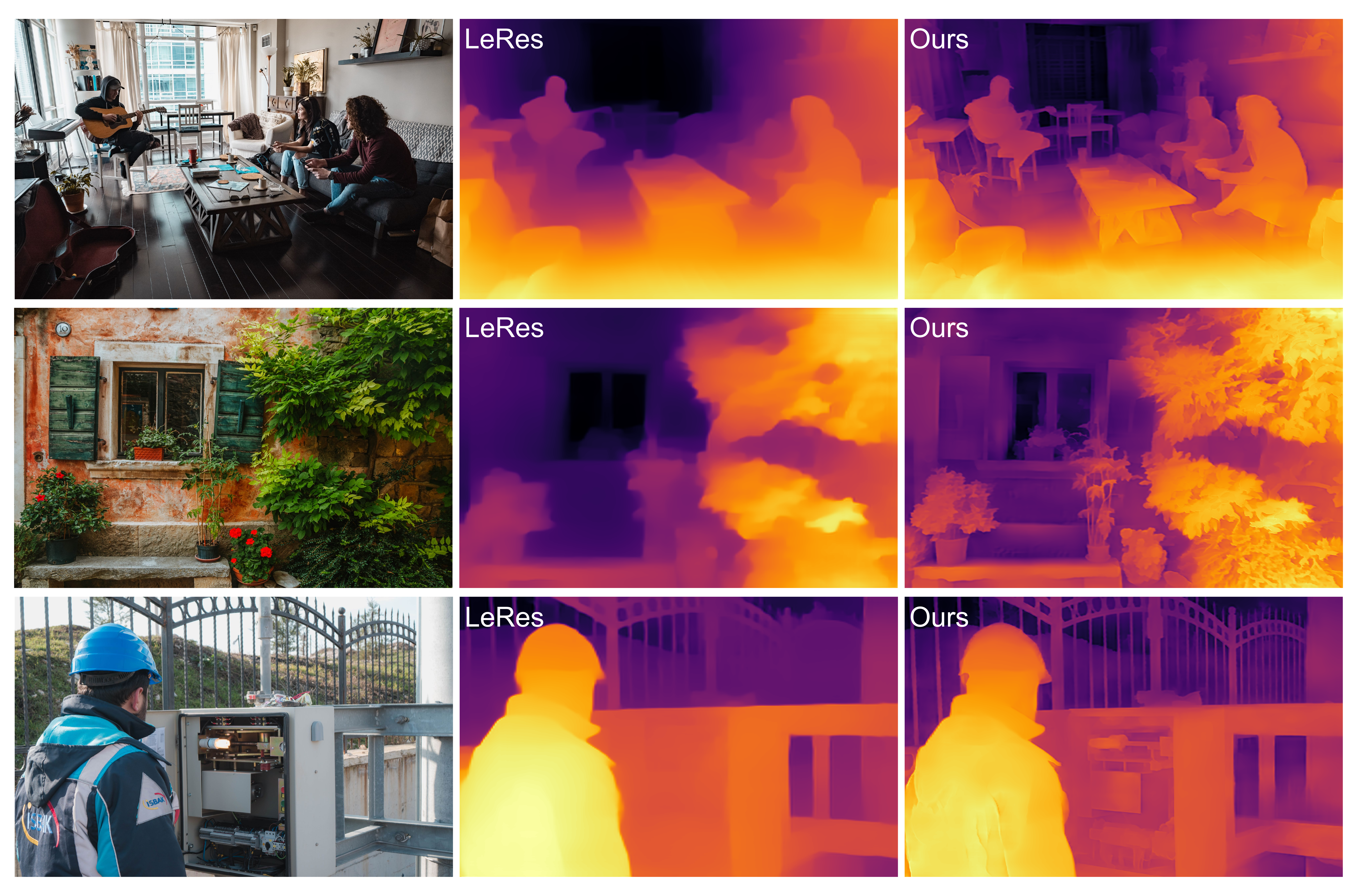}\
		\caption{We propose a multi-resolution depth map fusion method to recover high-resolution depth maps. Compared with the  single-resolution depth estimation network, the proposed method recovers wide levels of details. 
		}
		\label{fig:teaser}
		\vspace{-15pt}
	\end{figure}
	
	\vspace{-3pt}
	\vspace{-3pt}
\section{Introduction}
\label{sec:intro}

Depth is an essential information in a wide range of 3D vision applications, bridging 2D images to 3D world. Prior methods of image-based depth estimation rely on multi-view geometry or other scene priors and constraints. In real-life scenarios, multi-view images or additional inputs are not always accessible and monocular depth estimation (depth from a single image) is the most common case. Estimating depth from a single image is a challenging and ill-posed problem, where deep learning demonstrated great potential, by which priors are automatically learnt from training data. 

\begin{figure}
	\centering
	\includegraphics[width=0.99\linewidth]{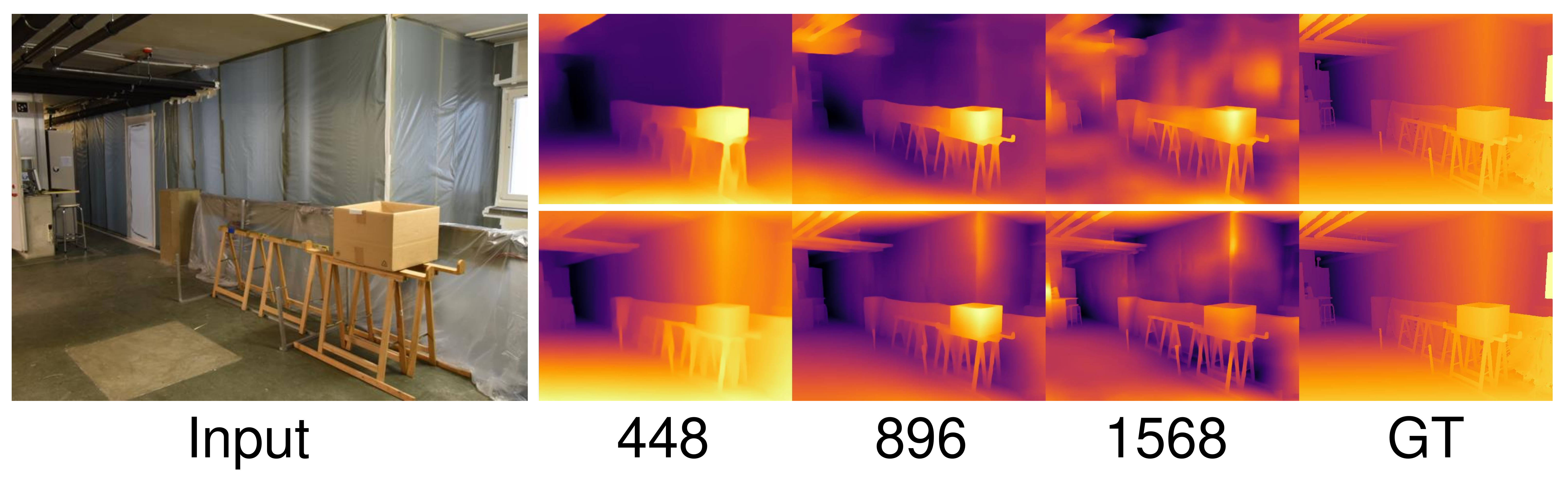}\
	\caption{ Different input sizes lead to different depth predictions. Low-resolution inputs recover more accurate depth values while higher-resolution inputs lead to more details in depth maps. The numbers denote the actual input sizes to the network after resizing. %For example, 448 denotes the input size of $448\times 448$. 
	GT denotes ground truth depth maps. The top row shows depth maps predicted by SGR~\cite{xian2020structure} and the bottom row shows depth maps predicted by LeRes~\cite{yin2021learning}. 
	}
	\label{fig:multires}
	\vspace{-15pt}
\end{figure}

As many other vision tasks based on deep-learning, the predictions of networks are more blurry than the input images, as shown in the second column ``$448$'' (denotes resolution of $448\times 448$) in Fig.~\ref{fig:multires}. 
%Details are losing due to \blue{due to the downsample of input image}the commonly used hourglass shape of network structures. 
Details are losing due to the down-sampling, max-pooling or convolution operations in the network. Furthermore, although fully convolutional networks take images at arbitrary resolutions as inputs, the networks are overfitted to the fixed resolution of training images. With a testing image at the training resolution, the estimated depth map is more accurate in values.  
%Using the pretrained input size of image
%one input image at a single resolution 
%would give an estimated depth map at a lower resolution.  
Considering the memory and training time, networks for monocular depth estimation are usually trained in a relatively low resolution. In two most recent monocular depth estimation methods SGR~\cite{xian2020structure} and LeRes~\cite{yin2021learning}, training images are at the resolution of $448\times 448$. Testing images at training low-resolution lead to more accurate depth predictions than those at different resolutions, as demonstrated in Fig.~\ref{fig:multires}. %Luckily, fully convolutional networks should be able to handle input images of arbitrary sizes. Although the result of a low-resolution testing image is more accurate in depth values, results of higher-resolution images
Testing images at original high-resolutions lead to inaccurate predictions of depth values, but keep much more local details. 
It motivates our explorations in multi-resolution depth map fusion, to combine the depth values and details predicted at different resolutions. It is studied that visual perceptions in natural images include various visual ``levels''~\cite{hubel1962receptive}, while different visual ``levels'' should be considered at the same time to get an overall plausible result. 

Instead of treating low- and high-resolution depth predictions equally by symmetry feature encoders, we adopt the core idea of Poisson fusion and propose a novel gradient-based composition network, fusing the gradient domain at high resolution into the depth map at low resolution.  

Poisson fusion is not fully differentiable and requires an additional manually labeled mask as input. In order to achieve an end-to-end automatic and differentiable pipeline, we propose a network conceptually inspired by Poisson fusion, without requiring a fusion mask. Based on the observation of higher value accuracy of low-resolution depth, and better texture details of high-resolution depth, we fuse the values of low-resolution depth and gradients of higher-resolution depth by guided filter~\cite{he2010guided}, which is a gradient-preserving filter, to get rid of the requirement of fusion masks. 
In guided filters, there are two parameters, the window size $r$ is set to adjust the receptive field, along with an edge threshold $\epsilon$. 
%Both parameters have to be tuned for different images, which is inpractical in automatic pipelines. 

Here, we fix both parameters for the whole dataset and select high quality data as training set to get a reasonable guided filtered depth map. This depth map is used as the supervision of the gradient domain. By adopting the guided filter in the training phase, our model learn to preserve the detail automatically without the help of guided filters while testing. %The guided filtering is used in the training phase only, it is not needed for the testing phase, avoiding parameters setting while testing. %We take the gradient domain of the guided filtered depth as a self-supervision in training, along with another self-supervised loss constraining the value consistency between the fused result and the low-resolution depth. 

\begin{figure*}
	\centering
	\includegraphics[width=\linewidth]{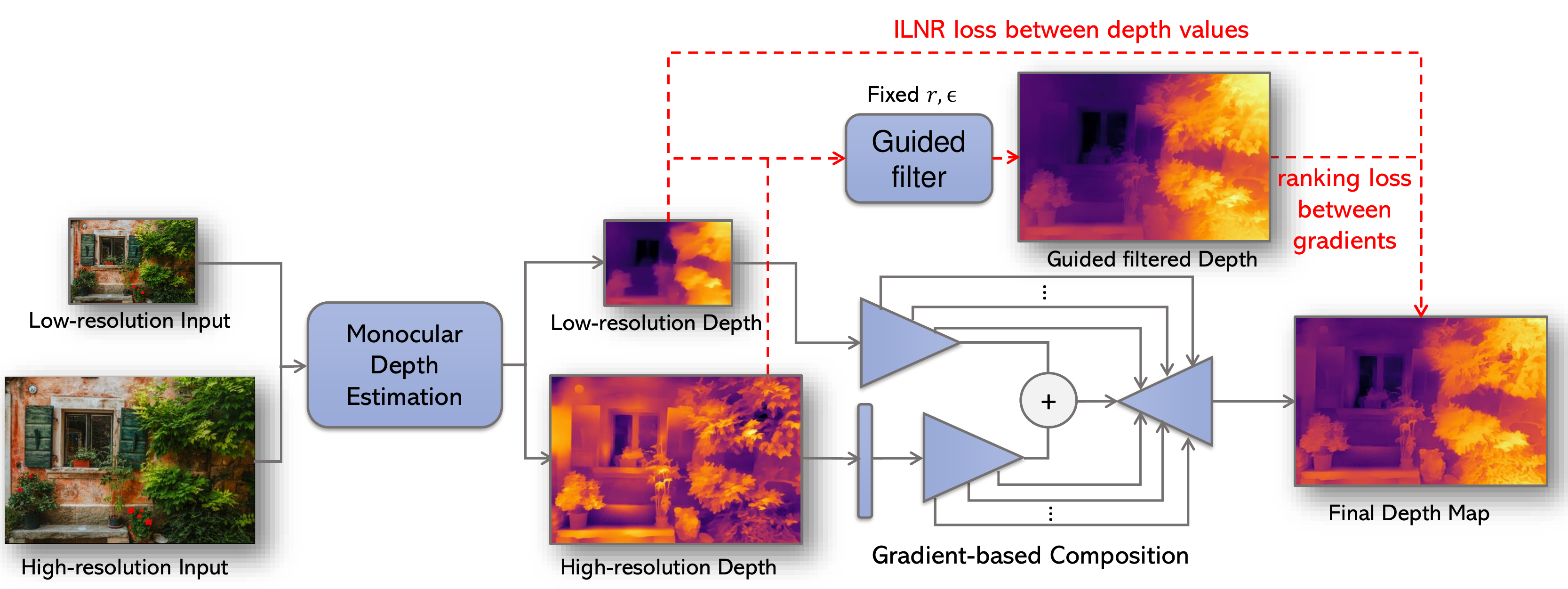}\
	\caption{Pipeline of the self-supervised multi-resolution depth map fusion method. With input images at low- and high-resolution, depth maps in two resolutions are predicted by the backbone monocular depth estimation network. The proposed gradient-based composition fuses two depth maps into a plausible one, training in a self-supervised fashion by a ILNR loss in the depth domain and a ranking loss in the gradients domain, supervised by the low-resolution depth map and the guided filtered depth map respectively. The end-to-end pipeline can be integrated with most fully convolutional backbone networks. Red dashed lines denote the procedures included in training phase only. }
	\label{fig:pipeline}
	\vspace{-15pt}
\end{figure*}

In details, the self-supervision is constrained by two separate losses, an image level normalized regression loss (ILNR loss) at the depth domain between the low-resolution depth map and the network prediction, and a novel ranking loss at the gradient domain between guided filtered depth map and the network prediction, as illustrated in Fig.~\ref{fig:pipeline}. By this self-supervised framework, no labeled data is needed in training. %\red{The training data of our network can be any images collected from any datasets or even from the Internet (images at high resolutions are preferable), no labeled data or any depth map ground truth is required.  }
 %In details, the self-training losses are similarities between the fused depth map and the low-resolution depth map, and a ranking loss between the gradients of fused depth map and the gradients of the guided filtered depth map. 
 
With most fully convolutional monocular depth estimation methods as backbone, our method effectively enhances their depth maps as in Fig.~\ref{fig:teaser}. Details are very well recovered, with the original depth accuracy preserved. %The training of our depth map fusion pipeline is self-supervised driven by consistency losses between the values of the fused result and input low-resolution depth, and the gradients of the fused result and input higher-resolution depth. 
%\red{add guided filtering into intro.} 
Our detailed pipeline is described in Fig.~\ref{fig:pipeline}. 
%While training, we first get the low- and high-resolution depth maps predicted by any fully convolutional backbone monocular depth estimation networks, then both depth maps are fed into guided filter with fixed parameters. 
%from the multi-resolution depth maps predicted by most backbone monocular depth estimation methods, we adopt a pyramid-style framework to fuse depths of three resolutions in three-step fusion. 
In the depth map fusion network (the gradient-based composition module in Fig.~\ref{fig:pipeline}), we firstly use a 1-layer convolution to get the approximate gradient map of the depth of higher resolution, then the depth map of lower resolution and the approximated gradient map of high-resolution input are fused in each layer of the 10-layer network to reconstruct the fused depth map. The network structure is designed specifically for a gradient-based composition inspired by Poisson fusion. 
%The self-supervision is driven by the consistency with depth values of low-resolution depth maps, and depth gradients of guided filtered depth maps. Note the guided filtered depth map is only needed for training, and while testing, one pass of the depth map fusion network is enough, no guided filtering is needed anymore.  %The fusion is repeated for three times, to fuse the low-, median- and high-resolution depth maps step by step.
%At last, the resulting depth map recovers a wide range of details while preserving the overall depth value accuracy. 
%As illustrated in Figure~\ref{fig:pipeline}, 
%Thus we propose a depth map fusion network,  training from self-supervised losses constraining the value of final depth map being similar to the low-resolution depth map and the gradient being similar to the guided filtered depth. It is very similar to Poisson fusion. 

Miangoleh et al.~\cite{miangoleh2021boosting} describe similar observations and 
%propose a two-resolution depth merging method. They
adopt GAN to merge low- and high-resolution depth maps of selected patches. Their method (BMD) effectively enhance the details in final depth maps but image noises would confuse their method to decide the high- and low-frequency patches. BMD is also time-consuming with the iterative fashion. Our solutions for multi-resolution depth fusion has a good robustness to image noises, runs in real-time, and fully self-supervised in training. In Sec.~\ref{sec:experiments}, we demonstrated that the performance of our method is more stable for different levels of noises, while the state-of-the-art depth map fusion method BMD~\cite{miangoleh2021boosting} degenerates when the noise variance is high. Our method benefits from the one-shot depth fusion by the lightweight network design, running at 5.4 fps while BMD~\cite{miangoleh2021boosting} running at 0.07 fps in the same environment. 
Comprehensive evaluations and a large amount of ablations are conducted, proving the effectiveness of every critical modules in the proposed method.

Our contributions are summarized as follows: 

%\vspace{-3pt}
\begin{itemize}
\item A portable network module is proposed to improve fully convolutional monocular depth estimation networks through a multi-resolution gradient-based fusion approach. Our method take advantages of the depth predictions of different resolution inputs, preserving the details while maintaining the overall accuracy.
\item A self-supervised framework is introduced to find the optimal fused depth prediction. No labeled data is required. % to enhance other state-of-the-art alternatives. 
\item The method has good robustness to various image noises, and runs in real-time, while state-of-the-art depth map fusion method degenerates significantly with noises increasing, and takes seconds for each data. 
%\item In evaluations, the proposed method significantly improve the depth prediction performance over the state-of-the-art method\cite{yin2021learning} with an improvement on $AbsRel$ by $12.7\%$ on the zero-shot public KITTI dataset\cite{geiger2012we}.  \red{update the numbers in this para.}

\end{itemize}

	\vspace{-3pt}
	\section{Related work}
	
	Monocular depth estimation is an essential step for many 3D vision applications, such as autonomous driving, simultaneous localization and mapping (SLAM~\cite{bailey2006simultaneous}) systems. Estimating depth from one single image is a challenging ill-posed problem, while traditional methods usually require multiple images to explore depth cues. For example, structure from motion~\cite{levinson2011towards} is based on the feature correspondences among multi-view images. With only one image, it is infeasible to solve the ambiguities. 
	
	For ill-posed problems, deep neural networks show a good superiority. 
	Deep-learning based methods can be categorized by supervision styles. A most straight-forward solution is supervised learning. Eigen et al.~\cite{eigen2014depth} proposes the first supervised work to solve monocular depth estimation, by defining Euclidean losses between predictions and ground truths to train a two-component network (global and local network). Mayer et al.~\cite{mayer2016large} solves scene flow in a supervised manner. Monocular depth estimation, along with optical flow estimation, are working as sub-problems of scene flow estimation. Recently, pretrained layers in ResNet~\cite{he2016deep} are widely used in monocular depth estimation networks~\cite{xian2018monocular,ranftl2019towards,yin2021learning} to speed up the training. Semi-supervised methods~\cite{smolyanskiy2018importance,kuznietsov2017semi,amiri2019semi} training from stereo pairs are proposed to soften the requirement of direct supervision. They estimate disparity between two stereo images, and define a consistency between one input image, and the re-rendered image by estimated inverse depth, disparity, camera pose and the other input. Kuznietsov et al.~\cite{kuznietsov2017semi} use sparse supervision from LIDAR data and incorporate with berHu norm~\cite{zwald2012berhu}. Following works based on LIDAR data~\cite{he2018wearable,wu2019spatial} propose similar semi-supervised training pipelines. Unsupervised methods~\cite{godard2019digging,casser2019depth,bozorgtabar2019syndemo,zhu2018scores} are mostly relying on the constraint of re-projections between neighboring frames in their training image sequences. By getting rid of supervisions, these methods suffer from many problems such as scale ambiguities and scale inconsistencies. 
	Other than different supervisions, different losses or constraints are adopted to better constrain the problem. Methods based on Berhu loss~\cite{heise2013pm,zhang2018joint}, conditional random fields~\cite{li2015depth,liu2015learning,wang2015towards} or generative adversarial networks (GAN)~\cite{feng2019sganvo,jung2017depth,gwn2018generative} are proposed. 
	
	A common issue of deep-learning methods is that many depth details are lost in network outputs. Depth maps are usually blurry with inaccurate details such as planes and object boundaries. This issue exists in many vision problems, such as image segmentation and edge detection. Deep-learning methods generate results much more blurry than non-CNN methods. Although replacing input images of higher resolutions generate predictions with higher resolutions, it leads to inaccurate depth values, as shown in Figure~\ref{fig:multires}. It motivates our work to fuse the depth maps of different resolutions to get an overall plausible one. Traditional image fusion methods such as Poisson fusion~\cite{perez2003poisson}, Alpha blending~\cite{porter1984compositing} require additional inputs such as alpha weights or masks, which usually require manual labeling. The proposed method automatically decides which regions from two-resolution depth maps have to be fused and how to fuse them. A recent depth map fusion method~\cite{miangoleh2021boosting} uses GAN to fuse the low- and high-resolution depths. Our method solves two-resolution depth fusion by self-supervised gradient-domain composition, achieving better robustness on image noises and real-time performance.

	\vspace{-3pt}
\section{Method}
\vspace{-3pt}
\subsection{Overview}\label{sec:overview}
Our key observation for monocular depth map fusion is that the local details can be preserved in the gradient domain of depth estimation from a high-resolution input, while the global value accuracy is better estimated with low-resolution input. In other words, a convolutional neural network (CNN) can focus on different levels of details when dealing with input images of different resolutions. Therefore, fusing the predictions of multi-resolution inputs is a straightforward choice to enhance the depth estimation. The goal of our method is finding the optimal fusion operation $\oplus$ for $d_{\text{high}}$ and $d_{\text{low}}$, which are depth predictions of the input image $I$ at high- and low-resolutions:

\begin{equation}\label{eq:oplus}
	f = d_{\text{high}}\oplus d_{\text{low}},
\end{equation}

\noindent where $f$ is the fused depth estimation with enhanced details.

Inspired by the Poisson fusion, the fusion operation $\oplus$ should transplant the gradient domain of $d_{\text{high}}$ to $d_{\text{low}}$ for detail-preserving. Thus, we formulate the optimization of depth fusion as below,

\begin{equation}\label{eq:overall}
	\min_\oplus \iint _\Omega | \triangledown f - \triangledown d_{\text{high}}|\partial \Omega + \iint _{I-\Omega} | f - d_{\text{low}}|\partial \Omega,
\end{equation}

\noindent where $\triangledown$ denotes the gradients of an image. Note that the optimization of $\oplus$ only focus on the gradient domain within $\Omega$ and value domain among other areas $I-\Omega$, where $\Omega$ is the area $d_{\text{high}}$ has better details than $d_{\text{low}}$. We propose a self-supervised neural network to find the fusion operation $\oplus$ based on Eq.~\ref{eq:overall}.

Note that the classic Poisson fusion is not differentiable while calculating the fused gradient around boundaries of fusion area $\Omega$. We first introduce a multi-level gradient-based fusion network module to approximate the Poisson fusion, which is described in Sec.~\ref{sec:multires}. Since we have no supervision of $\Omega$ to train this fusion module, we propose a self-supervised framework based on the supervision of guided filtering with a novel training loss, and details are in Sec.~\ref{sec:self-supervised}. Therefore, the fusion module in our pipeline is fully differentiable and capable of preserving gradient details of proper fusion area $\Omega$ while maintaining overall consistency and the training is fully self-supervised. 
Implementation details are provided in Sec.~\ref{sec:implementation}. 
At last, the evaluations Sec.~\ref{sec:experiments} demonstrates the superiority of the method on depth estimation accuracy and detail preservation over the state-of-the-art alternatives with better efficiency, and robustness to image noises and complicated textures.
\vspace{-3pt}
\subsection{Multi-level gradient-based depth fusion}\label{sec:multires}

\textbf{Monocular depth estimation.} Our multi-level gradient-based depth fusion module requires monocular depth estimation with different resolutions as inputs. LeRes\cite{yin2021learning} is a novel and state-of-the-art monocular depth estimation network that can provide good depth initials. However, it also lacks the capacity of providing clear depth details. We adopt LeRes\cite{yin2021learning} as the backbone to produce the depth 
initialization with different resolutions for depth fusion. As shown in Fig.~\ref{fig:multires}, the prediction of images with higher resolution contains more details while the lower resolution image can achieve higher overall accuracy. For each input $I$, we adopt 
two different resolutions $I_\text{low} ,I_\text{high}$ for the fusion operation, the corresponding predictions are $d_\text{low},d_\text{high}$ respectively. 

Note that our method can also work with other monocular depth estimation backbones. The evaluations with other three monocular single-resolution methods~\cite{xian2020structure,yuan2022newcrfs,ranftl2021vision} as backbone methods are provided in Sec.~\ref{sec:experiments}. 

\noindent \textbf{Differentiable gradient-domain composition.} Poisson fusion could be a good candidate for constructing the depth fusion module since it takes the gradient consistency into optimization. To ensure that the whole framework of our method can be trained in an end-to-end manner, we need to find a differentiable approximation to deal with the truncation of gradient backward around the merging boundaries.

To avoid the gradient truncation, we adopt an encoder-decoder framework to formulate the fusion module which can utilize $\Omega$ implicitly in the latent space. Then we can rewrite Eq.~\ref{eq:oplus} as below:

\begin{equation}\label{eq:en-decoder}
	f = \mathcal{D}(\mathcal{E}_l(d_\text{low})+\mathcal{E}_h(d_\text{high}),\Omega)
\end{equation}

\noindent where $\mathcal{D}$ is the decoder module, and $\mathcal{E}_l$, $\mathcal{E}_h$ are the encoder modules for low- and high-resolution depth estimations.

However, this formulation has two problems in implementation. First, $\Omega$ requires supervision but there are no such datasets. Thus, a self-supervised framework is proposed to solve this problem in Sec.~\ref{sec:self-supervised}. Second, training an encoder-decoder framework is highly inefficient if $\mathcal{E}_l$ and $\mathcal{E}_h$ do not share hyperparameters, however, sharing all hyperparameters will degrade the performance. Since we want to extract gradient information from $d_\text{high}$ and the depth values from $d_\text{low}$, it is natural to formulate $\mathcal{E}_h$ based on $\mathcal{E}_l$ with an additional one-level convolution layer $\mathcal{E}_g$ to extract gradients. % to extract gradients. 
\vspace{-3pt}
\begin{equation}\label{eq:gradient}
	\mathcal{E}_h(\cdot ) = \mathcal{E}_l(\mathcal{E}_g(\cdot ))
\end{equation}

$\mathcal{E}_g(d_\text{high})$ can be considered as a varied approximation with tunable parameters of $\triangledown d_{high}$. Therefore, Eq.~\ref{eq:en-decoder} can be rewritten as,
\vspace{-3pt}
\begin{equation}\label{eq:en-decoder_gradient}
	f = \mathcal{D}(\mathcal{E}_l(d_\text{low})+\mathcal{E}_l(\mathcal{E}_g(d_\text{high})),\Omega)
\end{equation}

\begin{figure}
	\centering
	\includegraphics[width=\linewidth]{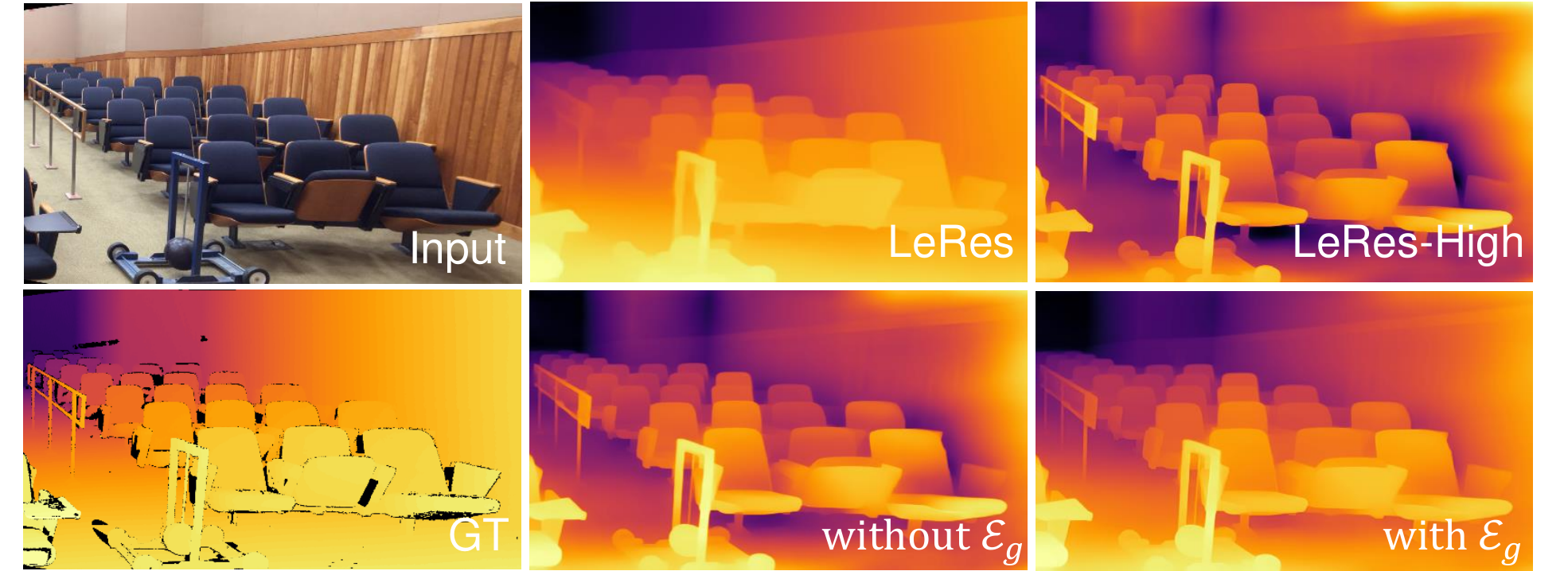}\
	\caption{Visual comparison upon $\mathcal{E}_g$. LeRes and LeRes-High are depth estimation results given by LeRes\cite{yin2021learning} with low- and high resolution inputs.}
	\label{fig:vis_g_layer}
	\vspace{-15pt}
\end{figure}

This solution is simple yet effective. Sharing most hyperparameters between $\mathcal{E}_l$ and $\mathcal{E}_h$ is highly efficient in the training stage. Our evaluation also demonstrates that the fusion performance will drop significantly if $\mathcal{E}_g$ is absent in the Eq.~\ref{eq:en-decoder_gradient}, which means this design is critical and essential. A visual comparison upon $\mathcal{E}_g$ is presented in Fig.~\ref{fig:vis_g_layer} which support our claimant as well.

\noindent \textbf{Multi-level fusion framework.}  To fully utilize $d_\text{low}$ and $d_\text{high}$ with neural networks, a simple one-step fusion with Eq.~\ref{eq:en-decoder_gradient} is not enough. A pyramid-style framework is introduced to fuse depth at different resolutions. We formulate Eq.~\ref{eq:en-decoder_gradient} with the multi-level encoder $\mathcal{E}^{2^i}_l$ as below:
\vspace{-3pt}
\begin{equation}\label{eq:encoder_multi_level}
	f = \mathcal{D}(\sum_{i=2}^{11}\mathcal{E}^{2^i}_l(d_\text{low})+\mathcal{E}^{2^i}_l(\mathcal{E}_g(d_\text{high})),\Omega)
\end{equation}

\noindent where $\mathcal{E}^{2^i}_l$ is multi-layer fully convolution module with resolution $2^i$. For implementation of Eq.~\ref{eq:encoder_multi_level}, we take $d_\text{low}$ and $\mathcal{E}_g(d_\text{high})$ into a multi-level convolution-based encoder individually. The convoluted outputs of $d_\text{low}$ and $\mathcal{E}_g(d_\text{high})$ from each level are then skip connected and be supplied as input for layered upsampling and convolution modules to reconstruct the final depth map.
\vspace{-3pt}
\subsection{Self-supervised framework of depth fusion}\label{sec:self-supervised}

\noindent \textbf{Self-supervision with guided filtering.} Image fusion based on Poisson equations introduces us to an interesting idea to fuse predictions of different resolution images. However, the classic Poissons-based fusion requires the manual labeling of the fusion area $\Omega$. To get rid of the manual steps we still need a proper $\Omega$ for depth fusion, training under the supervision of existing datasets is the most straightforward way. Unfortunately, no datasets are available for providing such information. We introduce a self-supervision mechanism driven by guided filtering to deal with this problem.

The guided filtering is an edge-preserving filtering, it has the gradient smoothing property. This filter fuses $d_\text{low}$ and $d_\text{high}$ together while keeping the gradient details of $d_\text{high}$ without manual mask labels like $\Omega$. This character makes it a perfect supervision for training our network. However, the guided filtering requires additional parameters to control fusion quality, and the tunable parameters are data dependent. Fortunately, we find that in the gradient domain of the fused image, a preset parameter works for the training set. Therefore, we can rewrite Eq.~\ref{eq:overall} as below,
\vspace{-3pt}
\begin{equation}\label{eq:gf_obj}
	\min_\oplus \iint | \triangledown f - \triangledown d_{\text{gf}}| + | f - d_{\text{low}}|\partial \Omega,
\end{equation}

\noindent where $d_{\text{gf}}$ is the fused result of $d_\text{low}$ and $d_\text{high}$ through guided filtering with a set of fixed parameters. The ablations in Section~\ref{sec:experiments} demonstrate that $d_{\text{gf}}$ provide effective supervision for training $\oplus$. Our fused module outperforms guided filtering as well in evaluations.

\begin{figure}
	\centering
	\includegraphics[width=\linewidth]{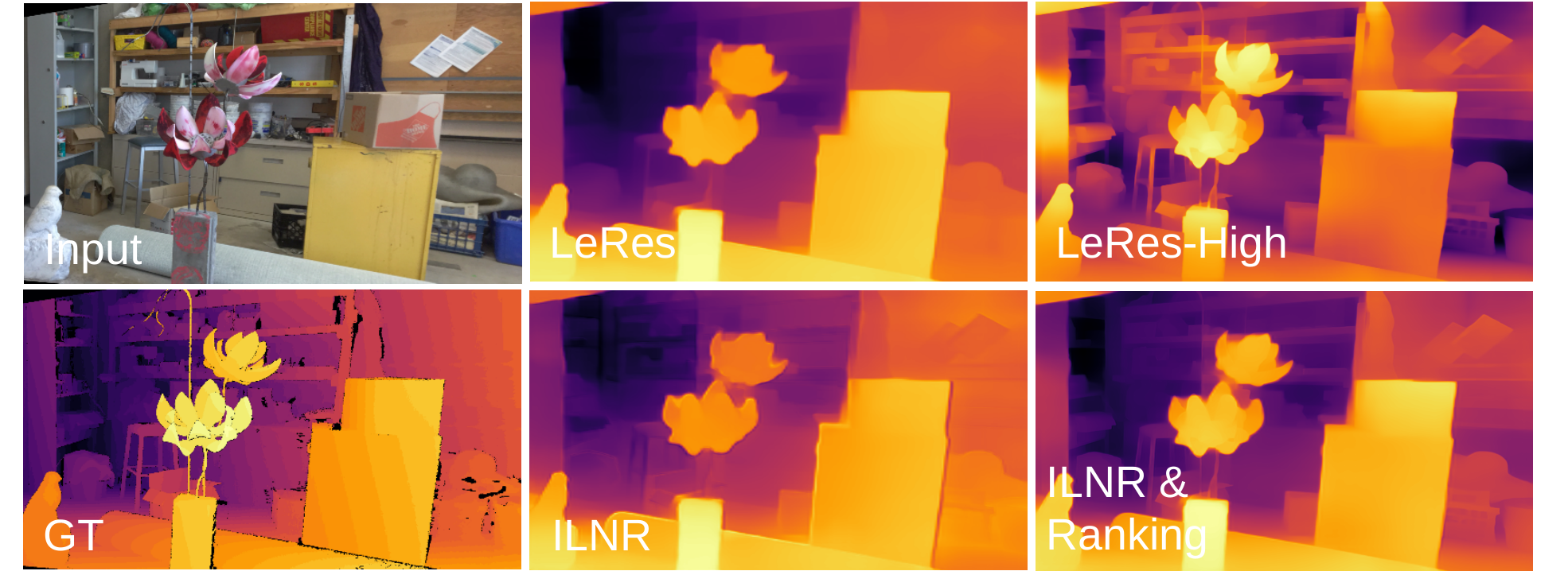}\
	\caption{Visual comparison upon $l_{\text{mILNR}}$ and $l_{\text{rank}}$. LeRes and LeRes-High are depth results by LeRes\cite{yin2021learning} with low- and high-resolution inputs. ILNR is the fused result training only by $l_{\text{mILNR}}$ while ours adopts both.}
	\label{fig:vis_losses}
	\vspace{-15pt}
\end{figure}

\noindent \textbf{Self-supervised loss.} The training objective Eq.~\ref{eq:gf_obj} is then optimized through the self-supervised loss $l_{\text{fusion}}$ as below,
\vspace{-3pt}
\begin{equation}\label{eq:fusion}
	l_{\text{fusion}}(f, d_{\text{gf}}, d_{\text{low}}) =  l_{\text{mILNR}}(f, d_{\text{low}}) + l_{\text{rank}}(f, d_{\text{gf}}),
\end{equation}

\noindent where $l_{\text{mILNR}}= \sum_{r}l_{\text{ILNR}}(f^r, d^r_{\text{low}})$ is multi-resolution image-level normalized regression (ILNR) loss~\cite{yin2021learning} which constrains the value domain of fused result $f$ to be similar as $d_{\text{low}}$ at every resolution $r$ levels. Another term $l_{\text{rank}}$ is a novel ranking loss inspired by \cite{xian2020structure} which constrains the gradient domain of $f$ being close to $d_{\text{gf}}$. Given a pair of points $i,j$, ${p^i_f\in f,p^i_{\text{gf}}\in d_{\text{gf}}}$ are pixels on $f$ and $d_{\text{gf}}$ respectively. $l_{\text{rank}} = \frac{1}{N} \sum_{i,j}E(p^i_f,p^j_f,p^i_{\text{gf}},p^j_{\text{gf}})$ is formulated based on these sampled pixel pairs, and $N$ is the number of pixel pairs. Point pair $i,j$ are sampled based on edge areas $M$ extracted from $\triangledown d_{\text{gf}}$ and $\triangledown d_{\text{high}}$ by Canny detection. %Details are in Sec. 2.3 of the Supplementary.

\vspace{-3pt}
\begin{equation}\label{eq:losses}
\scalebox{0.8}{	$E(p^i_f,p^j_f,p^i_{\text{gf}},p^j_{\text{gf}}) =
\left\{
             \begin{array}{lr}
			  \log(1+e^{-\frac{1}{|(p^i_f-p^j_f)-(p^i_{\text{gf}}-p^j_{\text{gf}})+\sigma|}}),&  z_{ij}=1,\\
			  |p^i_f-p^j_f|^2, & z_{ij}=0,
             \end{array}
\right.$
}\end{equation}

\noindent where $z_{ij}$ is an indicator, $z_{ij}=1$ means pixel $i,j$ are located on different sides of an edge of $M$ while $z_{ij}=0$ means they located on the same side. $\sigma$ is a regular term
 for robust compuatation. A visual comparison upon our loss terms $l_{\text{mILNR}}$ and $l_{\text{rank}}$ is presented in Fig.~\ref{fig:vis_losses}, which demonstrate the our design of losses is effective.

%Implementation details are included in the Supplementary. 
	
	\vspace{-5pt}
\section{Experiments}\label{sec:experiments}
\begin{table*}
	\centering
	\scalebox{0.8}{
		%\caption{\red{replace tab.1, add the result of the backbone}}
		\label{tab}
\begin{tabular}{c|ccccc|ccccc|ccccc} 
\hline
\multirow{2}{*}{Methods} & \multicolumn{5}{c|}{Multiscopic} & \multicolumn{5}{c|}{Middlebury2021} & \multicolumn{5}{c}{Hypersim} \\ 
\cline{2-16}
 & SqRel$\downarrow$ & rms$\downarrow$ & log10$\downarrow$ & $\delta_1\uparrow$ & $\delta_2\uparrow$ & SqRel$\downarrow$ & rms$\downarrow$ & log10$\downarrow$ & $\delta_1\uparrow$ & $\delta_2\uparrow$ & SqRel$\downarrow$ & rms$\downarrow$ & log10$\downarrow$ & $\delta_1\uparrow$ & $\delta_2\uparrow$ \\ 
\hline
SGR & 9.161 & 14.031 & 0.086 & 0.745 & 0.904 & 0.846 & 3.948 & 0.067 & 0.773 & 0.94 & 0.593 & 1.536 & 0.102 & 0.612 & 0.854 \\
NeWCRFs & 11.031 & 14.658 & 0.088 & 0.749 & 0.899 & 0.829 & 3.724 & 0.058 & 0.830 & 0.952 & 0.513 & 1.322 & 0.088 & 0.694 & 0.881 \\
DPT & \uline{4.021} & \uline{9.781} & \uline{0.059} & \textbf{0.841} & 0.904 & 0.700 & 3.698 & 0.060 & 0.827 & 0.956 & 0.327 & 1.145 & 0.083 & 0.734 & 0.914 \\
LeRes & 9.168 & 13.12 & 0.082 & 0.776 & 0.909 & 0.464 & 3.042 & \uline{0.052} & 0.847 & \textbf{0.969} & 0.319 & 1.011 & \textbf{0.071} & 0.768 & \textbf{0.924} \\ 
\hline
SGR-GF & 9.314 & 14.107 & 0.087 & 0.743 & 0.903 & 0.844 & 3.9 & 0.067 & 0.773 & 0.94 & 0.600 & 1.539 & 0.103 & 0.612 & 0.854 \\
NeWCRFs-GF & 10.601 & 14.407 & 0.087 & 0.751 & 0.901 & 0.796 & 3.549 & 0.058 & 0.832 & 0.953 & 0.518 & 1.320 & 0.088 & 0.694 & 0.881 \\
DPT-GF & 4.142 & 10.060 & 0.060 & \uline{0.835} & \uline{0.935} & 0.685 & 3.653 & 0.060 & 0.825 & 0.956 & 0.332 & 1.149 & 0.083 & 0.733 & 0.914 \\
LeRes-GF & 9.01 & 13.063 & 0.082 & 0.776 & 0.91 & \uline{0.457} & \uline{2.976} & \uline{0.052} & \uline{0.849} & \uline{0.968} & 0.324 & 1.011 & \uline{0.072} & \uline{0.769} & 0.922 \\ 
\hline
3DK & 9.379 & 14.879 & 0.077 & 0.73 & 0.895 & 0.911 & 4.18 & 0.069 & 0.745 & 0.945 & 0.718 & 1.521 & 0.108 & 0.610 & 0.840 \\
LeRes-BMD & 9.259 & 13.101 & 0.083 & 0.773 & 0.91 & 0.487 & 3.014 & 0.055 & 0.844 & 0.96 & \uline{0.312} & \textbf{0.993} & \uline{0.072} & \uline{0.769} & 0.922 \\ 
\hline
SGR-Ours & 9.144 & 14.0 & 0.086 & 0.746 & 0.905 & 0.816 & 3.858 & 0.067 & 0.776 & 0.942 & 0.605 & 1.549 & 0.103 & 0.609 & 0.851 \\
NeWCRFs-Ours & 10.405 & 14.299 & 0.087 & 0.752 & 0.902 & 0.786 & 3.587 & 0.057 & 0.832 & 0.953 & 0.520 & 1.324 & 0.089 & 0.687 & 0.880 \\
DPT-Ours & \textbf{3.998} & \textbf{9.759} & \textbf{0.058} & \textbf{0.841} & \textbf{0.938} & 0.647 & 3.533 & 0.059 & 0.832 & 0.958 & \textbf{0.311} & 1.109 & 0.081 & 0.745 & 0.915 \\
LeRes-Ours & 8.833 & 12.921 & 0.081 & 0.781 & 0.911 & \textbf{0.444} & \textbf{2.963} & \textbf{0.051} & \textbf{0.853} & \textbf{0.969} & 0.315 & \uline{0.999} & \textbf{0.071} & \textbf{0.77} & \uline{0.923} \\
\hline
	\end{tabular}}
	\caption{The quantitative evaluations on three benchmark datasets. Bold numbers denote the best result while underlined numbers are second best. Depth map fusion methods on monocular depth estimation backbones are presented as ``backbone-fusion method''. Our method achieve the best performance on $13/15$ metrics in these 3 datasets.} \label{table:quantitative}
	\vspace{-5pt}
\end{table*}

In Sec.~\ref{sec:dataset}, we introduce the benchmark datasets used in evaluations. In Sec.~\ref{sec:comparison}, we compare the proposed method with several state-of-the-art monocular depth estimation methods and depth refinement methods in aspects of several error metrics, robustness to noises, and running time. 
%, where our method outperforms both quantitatively and qualitatively in aspects of .
Lastly, in Sec.~\ref{sec:ablations}, we conduct several ablations to study the effectiveness of several critical design in the proposed pipeline. %Additional experiments and codes are in the Supplementary. 
\vspace{-3pt}
\subsection{Benchmark datasets and evaluation metrics}\label{sec:dataset}
\vspace{-1pt}
To evaluate the depth estimation ability of our method, we adopt several commonly used zero-shot datasets, which are Multiscopic~\cite{yuan2021stereo},  Middlebury2021~\cite{scharstein2014high} and Hypersim~\cite{roberts:2021}. %Evaluations on NYU~\cite{Silberman:ECCV12} and KITTI~\cite{Uhrig2017THREEDV} datasets are in the Supplementary.
% and Diode~\cite{vasiljevic2019dode}. 
All benchmark datasets are unseen during training. %, and their detailed descriptions can be found in the Supplementary. 

%Diode (Dense Indoor and Outdoor DEpth)~\cite{vasiljevic2019dode} is a dataset of high-resolution RGB-D images of various indoor and outdoor scenes. 

%In our evaluations, for NYU dataset we test on the testing set of 654 images. For KITTI dataset, the released validation dataset of 1000 images are used for evaluation. For DIODE, we use their validation dataset of 771 images including both indoor and outdoor data. For Sintel dataset, we sample 15 sequences containing 740 images. For Ibims-1, we test on the whole dataset of 100 images. 

In evaluations, we test on the whole set of Middlebury2021~\cite{scharstein2014high}, including 24 real scenes. For Multiscopic~\cite{yuan2021stereo} dataset, we evaluate on synthetic test data, containing 100 high resolution indoor scenes. For Hypersim~\cite{roberts:2021}, we evaluate on three subsets of 286 tone-mapped images generated by the released codes. On these datasets, we evaluate several error metrics. $SqRel$ and $rms$ denote two common metrics, which are the square relative error and the root mean square error, respectively. 
%Definitions of the square relative error $SqRel$ and the root mean square error $rms$ are included in the Supplementary. 
The mean absolute logarithmic error $log_{10}$ is defined as 
$log_{10}=\frac{1}{N}\sum {\left \|log(d_{i}^{*})-log(d_{i})\right \|}$. The error metric $delta$ describes the percentage of pixels satisfying $\delta =max(\frac{d_{i}^{*}}{d_{i}},\frac{d_{i}}{d_{i}^{*}})<1.25^{k}$ (i.e. $\delta_{1.25}^{k}$), $k = 1, 2$ are adopted to evaluate our performance, which are $\delta_1, \delta_2$ in the tables. $D^3R$ metric measures the detected edge error, which is introduced in \cite{miangoleh2021boosting}. 
In all metrics $d_{i}^{*}$ is the ground truth depth value and $d_{i}$ denote the predict value at pixel $i$. Due to the scale ambiguity in monocular depth estimation, we follow Ranftl et al.~\shortcite{ranftl2019towards} to align the scale and shift using the least squares before computing errors. %\red{More details as well as a comparison of these benchmark datasets are summarized in Table~\ref{table:dataset}. }

% \begin{table}[t]
% 	\centering 
% 	\scalebox{0.7}{
% 		\begin{tabular}{c c c c c} 
% 			\hline 
% 			Datasets&$\#$scenes&Scene type&Metrics&Data type\\
% 			\hline
% 			Middlebury2021&24&Indoor&rel, rms, $\delta_{1.25}$&Structured-light\\
% 			\hline
% 			Multiscopic&100&Indoor&rel, rms, $\delta_{1.25}$&Synthetic\\
% 			\hline
% 		\end{tabular}
% 	}
% 	\caption{Details of benchmark datasets used in evaluations. }\label{table:dataset} 
% \end{table}

%we conduct several experiments to demonstrate the effectiveness of our method, including a comparison with state-of-the-art methods, a comparison of with or without our proposed guide mask, and a comparison of simplified network architecture. We also compared our method with the existing 融合？ method. 
\vspace{-3pt}
\subsection{Comparisons to state-of-the-arts}\label{sec:comparison}

\noindent \textbf{Quantitative Evaluation.} We compare our method with two state-of-the-art fusion based monocular depth estimation alternatives, BMD~\cite{miangoleh2021boosting}, and 3DK~\cite{niklaus20193d}, in Tab.~\ref{table:quantitative}. To demonstrate that our self-supervised framework is effective, we also present the performance of monocular depth estimation methods applying with guided filter~\cite{he2010guided} as a baseline. In the comparison, we adopt the trained models released by the authors and evaluate with their default configuration. 

%When computing errors, we follow the processing pipeline of \cite{ranftl2019towards} and compare scale-invariant errors, since monocular depth estimation can only predict relative depth values. The performances on seven error metrics are shown in Table~\ref{table:quantitative}. While higher  $\delta_{1.25}$ is better, other metrics are the lower the better. 

Tab.~\ref{table:quantitative} demonstrate that our method outperform all the other fusion based depth estimation alternatives on most benchmarks and error metrics. Our method also produce better fusion results than guided filtering which is our training supervision. It means that our network takes advantages of both the low-resolution input and the guided filtered fusion result but not strictly constrained. %The evaluation results demonstrate the superiority of our method over other state-of-the-art methods.
It should be noticed that 
our method was trained with the monocular depth estimation backbone LeRes~\cite{yin2021learning}. The trained fusion module can be directly integrated with other fully convolutional monocular depth estimation networks such as SGR~\cite{xian2020structure}, NeWCRFs~\cite{yuan2022newcrfs} and DPT~\cite{ranftl2021vision}, without any fine-tuning. The proposed method is portable and can be easily incorporated into many state-of-the-art depth estimation models. 

We also evaluate the details recovered by our method. We present the comparison between our method and two backbone methods with $D^3R$ metric in Tab.~\ref{tab:D^3R}, which measure the edge correctness of the estimation depth details.
\begin{comment}

\begin{table}
	\centering
	\scalebox{0.8}{
	\begin{tabular}{p{2.5cm}<{\centering}|p{2cm}<{\centering}p{2cm}<{\centering}} 
	\hline
	Method &  Multiscopic & Middlebury2021\\ 
	\hline
	LeRes & 0.570 & 0.719\\
	SGR & 0.576 & 0.735\\
	Ours & \textbf{0.542} & \textbf{0.684}\\
	\hline
	\end{tabular}}
	\caption{Evaluation of our fusion module on edge correctness of depth by $D^3R$ metric, where lower values are better. }\label{tab:D^3R}
	\vspace{-5pt}
	\end{table}
\end{comment}
\begin{table}
	\centering
	\scalebox{0.8}{
	\begin{tabular}{p{2.5cm}<{\centering}|p{2cm}<{\centering}p{2cm}<{\centering}} 
	\hline
	Method &  Multiscopic & Middlebury2021\\ 
	\hline
	SGR & 0.576 & 0.735\\
	DPT & 0.594 & 0.613\\
	NeWCRFs & 0.767 & 0.737\\
	LeRes & 0.570 & 0.719\\
	Ours & \textbf{0.542} & \textbf{0.589}\\
	\hline
	\end{tabular}}
	\caption{Evaluation of our fusion module on edge correctness of depth by $D^3R$ metric, where lower values are better. }\label{tab:D^3R}
	\vspace{-15pt}
	\end{table}
%We finally evaluate our multi-resolution depth fusion framework on two monocular depth estimation methods, SGR~\cite{xian2020structure} and  LeRes~\cite{yin2021learning}. We also compare the quality of our multi-resolution fusion network with several depth refinement methods including BMD~\cite{miangoleh2021boosting}, 3DK~\cite{niklaus20193d} and a baseline method guided filter~\cite{he2010guided}.
\begin{figure}
	\centering
	\includegraphics[width=\linewidth]{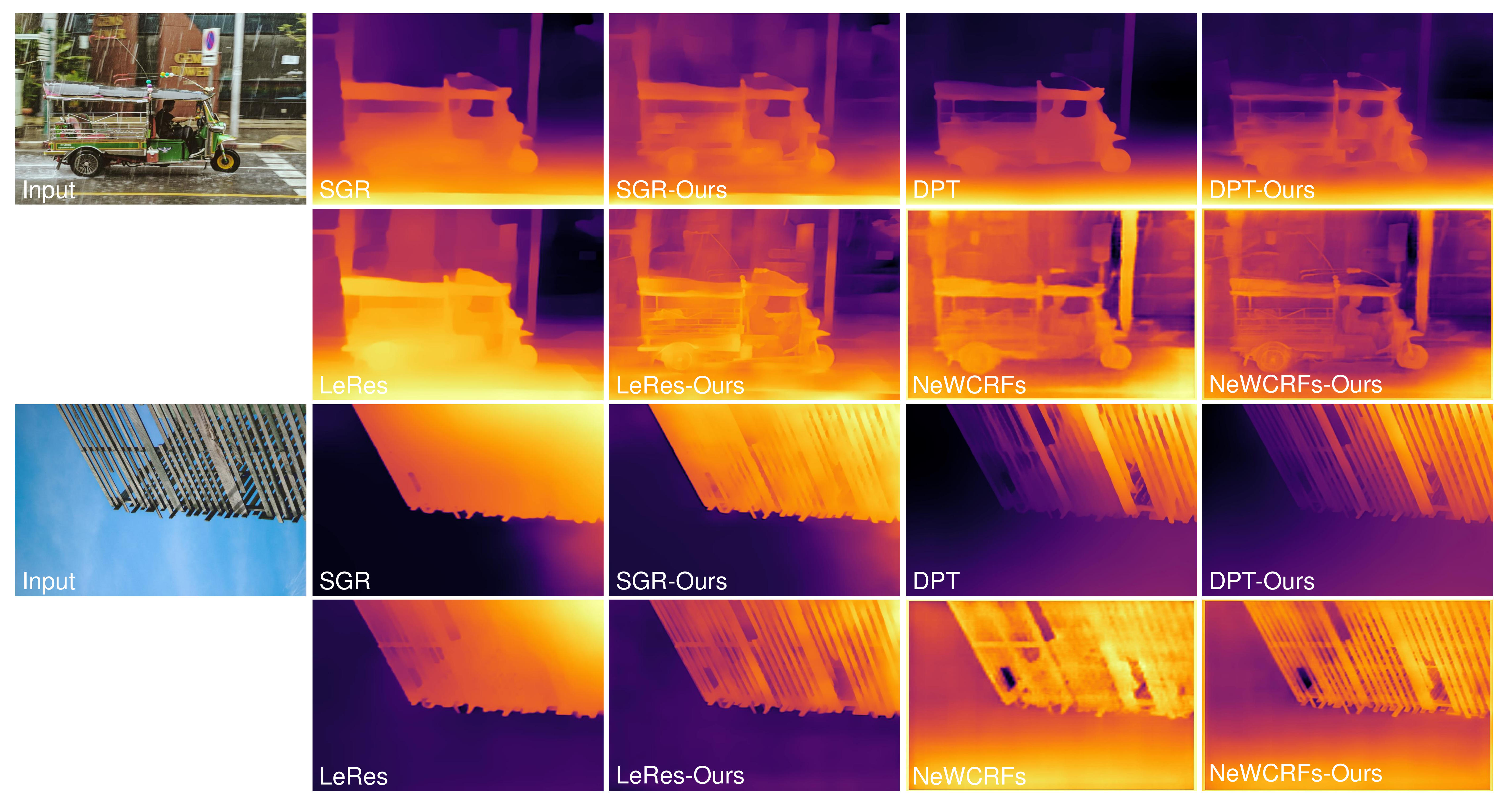}\
	\caption{Qualitative comparisons on unseen natural images from the Internet. Our method successfully boosts the performances of the backbone monocular depth estimation methods, and recovers the details in depth maps. }
	\label{fig:widecomparison}
	\vspace{-15pt}
\end{figure}

\noindent \textbf{Visual Comparisons.} The qualitative evaluations are presented in Fig.~\ref{fig:widecomparison}. As discussed earlier, most monocular depth estimation methods suffer from blurry predictions. The improvement of detail-preserving by our multi-resolution depth map fusion is significant. Most details missing in the backbone methods are successfully recovered, while keeping the original depth values correct. Additional examples are shown in Fig.~\ref{fig:visuMidd}-\ref{fig:visuWild_3}, our fusion method can significantly improve performance of SGR~\cite{xian2020structure} and LeRes~\cite{yin2021learning}. %In the video, we test on several Internet videos, and perform our monocular (single-image) method per frame. %On different backbone methods, the proposed method successfully recovers details in depth maps. The quantitative evaluations are listed in Table~\ref{table:quantitative}. We adopt several error metrics, which are commonly used for depth estimation. 
%where the superiority of our multi-resolution depth fusion are more obvious. 
%fitting the scenario of the proposed multi-resolution framework. 

\noindent \textbf{Anti-noise Evaluation.} One of the main advantages of our method over the state-of-the-art depth fusion method BMD~\cite{miangoleh2021boosting} is the noise robustness enabled by our gradient-based fusion. Since BMD determines fusion areas explicitly by edge detection, its performance will drop significantly while the input images include noises. We compare the anti-noise ability of our methods with BMD. For evaluation, we add Gaussian noises or Pepper noises to the input image. The mean value of Gaussian noises is 0, the variances changes from 0.001 to 0.009 for Gaussian noises, and the signal-noise ratio changes from $100\%$ to $95\%$ for Pepper noises. Fig.~\ref{fig:noise_line} shows the variety of $\delta_{1}$ value when adopting different variance levels of Gaussian (left) or Pepper noises (right). Fig.~\ref{fig:pepper} also presents visual examples of our method comparing with BMD, on input images with two types of noises. %We use LeRes as the backbone of BMD and ours, $*$ in Fig.~\ref{fig:noise} denote the result with Gauss noise (variance is 0.006).

\begin{figure}[!t]
	\centering
	\includegraphics[width=1\linewidth]{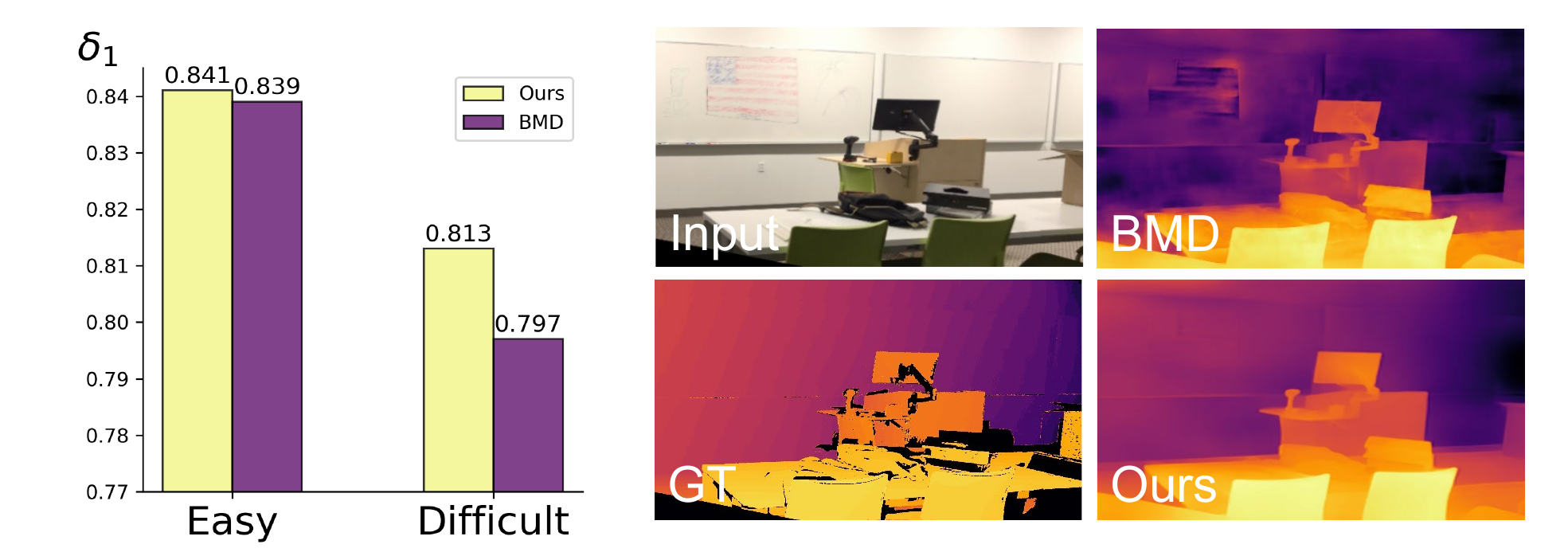}\
	\caption{Left: Quantitative evaluation between our method and BMD on easy and difficult datasets regarding to texture complexities. Ours outperform BMD more significantly on the difficult subset. Right: Visual comparisons between our method and BMD. 
		%Input image, depth estimation from low resolution input, depth estimation from high resolution input, ground truth depth, BMD fusion result, and our result are presented. 
	}
	\label{fig:texture_vis}
	\vspace{-5pt}
\end{figure}

\begin{figure}
	\centering
	\includegraphics[width=\linewidth]{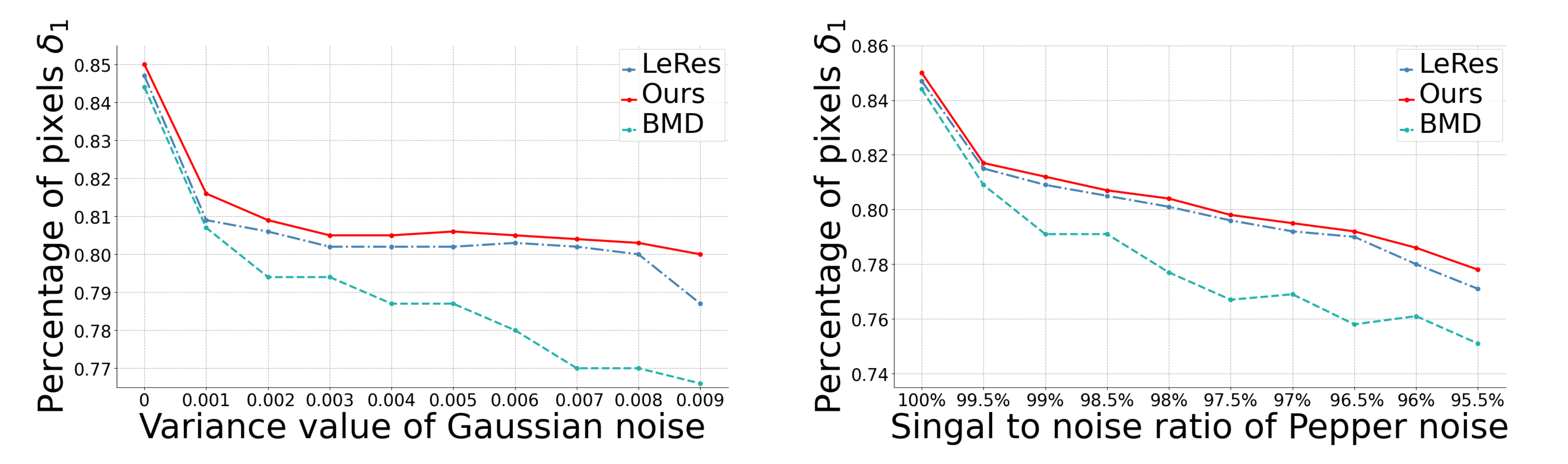}\
	\caption{Quantitative evaluation of anti-noise ability of LeRes, BMD and ours. The vertical axis denotes the value of $\delta_{1.25}$, which is the higher the better.}
	\label{fig:noise_line}
	\vspace{-5pt}
\end{figure}

Fusion based methods such as BMD can also be easily influenced by the complicated textures. We split Middlebury2021 benchmark into two sub-sets based on the edge number detected in the ground truth depth images. Less edges on depth maps means more depth independent textures exist. The plot at the left of Fig.~\ref{fig:texture_vis} demonstrates that our method outperforms BMD more significantly on the difficult sub-set.  At the right of Fig.~\ref{fig:texture_vis}, a visual comparison on a difficult example is shown, where BMD suffers from the complicated textures and extract many texture details into the depth map (e.g. paintings on the whiteboard).

\noindent \textbf{Running time.} Our fusion module is one-shot and requires no additional complex processing, it is highly efficient and only takes $68\%$ processing time comparing with LeRes for depth estimation of high resolution input. The fusion processing of our method is 10X faster than guided filter while present better performance. The whole pipeline of our method, including the depth estimation time of low- and high-resolution inputs, is more than 80X faster than BMD. 
Detailed time and computational statistics can be found in Tab.~\ref{tab:ruuningtime}. 
%Through the whole evaluation, we adopt LeRes~\cite{yin2021learning} as the backbone method of BMD. For guided filter parameters radius and regularization, we fixed at $\frac{W_{i}}{12}$ and $10^{-12}$ (${W_{i}}$ denote the width of the input image). And for the input resolution, we adopt $W_{b}$ (original size) for backbone to predict low-res depth while using $W_{b} * \frac{r}{2}$ to predict high-res depth ($r$ is the upscale parameter, through the experiments we fixed $r$ at 3). 

\begin{comment}

\begin{figure}
	\centering
	\includegraphics[width=0.99\linewidth]{noise-new.pdf}\
	\caption{Comparisons of our method with \cite{miangoleh2021boosting} on the same depth estimation backbone LeRes~\cite{yin2021learning} on images with Gaussian noises. $*$ denote the result from image with higher Gauss noise. It shows that our method is more robust to image noises, where BMD~\cite{miangoleh2021boosting} introduces more artifacts.  }
	\label{fig:noise}
\end{figure}

\end{comment}
\vspace{-3pt}
\subsection{Ablation studies}\label{sec:ablations}
\begin{table}[t]
	\centering
	\scalebox{0.65}{
		\arrayrulecolor{black}
		\begin{tabular}{c|cccc|c} 
			\hline
			\multirow{2}{*}{} & \multicolumn{4}{c|}{Loss function} & Result \\ 
			\cline{2-6}
			& ILNR & Gradient & SGR Ranking & Ours Ranking & error \\ 
			\cline{1-5}\arrayrulecolor{black}\cline{6-6}
			Setting A & Low-res depth & High-res depth & $\times$ & $\times$ & 0.734 \\
			Setting B & Low-res depth & $\times$ & High-res depth & $\times$ & 0.722 \\
			Setting C & Low-res depth & $\times$ & $\times$ & High-res depth & \uline{0.688} \\
			Setting D & Guided depth & Guided depth & $\times$ & $\times$ & 0.723 \\
			Setting E & Guided depth & $\times$ & Guided depth & $\times$ & 0.714 \\
			Setting F & Guided depth & $\times$ & $\times$ & Guided depth & 0.711 \\
			\arrayrulecolor{black}\hline
			Ours & Low-res depth & $\times$ & $\times$ & Guided depth & \textbf{0.684} \\
			\hline
	\end{tabular}}
	\caption{Ablation study on different training loss settings. %Here setting F is our final setting. 
		%and error calculation we follow Xian et al.\cite{xian2020structure}
	}
	\label{table:ablation} 
	%\vspace{-25pt}
\end{table}
We evaluate the effectiveness of critical designs in our method. All ablation alternatives are trained for 30 epochs, under identical training configurations.  %Our full performance after 150 epochs are in Table~\ref{table:quantitative}. 
%We adopt guided filter to get a enhanced fused depth, where the radius and regularization are fixed at $\frac{W_{i}}{12}$ and $1e-12$ (${W_{i}}$ (${W_{i}}$ denote the width of the input image).

We first investigate the effects on model performance of different type of loss functions, including ILNR loss, gradient loss~\cite{li2018megadepth} and ranking loss. We construct the ablation study with 7 different setting of training losses as shown in Tab.~\ref{table:ablation} on Middlebury2021. We adopt ILNR loss to supervise the value domain in our fusion network comparing with low resolution result and guided fused result, and adopt gradient loss\cite{li2018megadepth}, original ranking loss \cite{xian2020structure} or the proposed ranking loss to supervise the gradient domain comparing with high resolution depth and guided fused result. 
%In setting A, we use gradient loss\cite{} to learn the gradient from high-res backbone result while utilize ILNR loss to keep the value of low-res backbone result. As a contrast, we use guided filter to get a guided depth and learn the gradient and depth value from guided depth instead in setting D. In setting B, we use ILNR loss to learn the value of low-res depth while utilize the original ranking loss\cite{xian2020structure} to learn the detail from high-res depth. In setting F(our final design), we train the network with ILNR loss and our ranking loss, optimize the difference between fusion result and guided depth.
Tab.~\ref{table:ablation} shows that our configuration outperform other alternatives on $D^3R$ metric. %The visual examples presented in Fig. 3 of the Supplementary also demonstrate that the proposed loss functions and optimization targets lead to better performance for recovering the depth details. 
We also evaluate the validation of our differential gradient-domain composition design. We compare the depth estimation performance with and without $\mathcal{E}_g$. The results in Tab.~\ref{tab:e_g_ablation} demonstrate that this simple design is critical and can significantly improve the detail preserving performance. The visual comparison is presented in Fig.~\ref{fig:vis_g_layer}. % and more can be found in the Supplementary.

%Optimization target:

%In order to illustrate the effectiveness of our designed optimization target, the four settings can be seen as two comparison groups - Setting A with Setting C, Setting B with Setting D. Two comparison groups have the same adoption of loss function respectively, and the setting utilize guided depth has the lower error in each group. 

%Loss type:

%Similarly, the four settings can be seen as two comparison groups - Setting A with Setting B, Setting C with Setting D, to demonstrate the effectiveness of our ranking loss.  The settings in each group use ILNR loss to preserve the depth, while using ranking loss improve the ability of recovering the detail. Quantitative result and qualitative result improvements can be easily find in Fig~\ref{fig:loss} and  Tab~\ref{table:ablation}

\begin{table}
\centering
\scalebox{0.8}{
\begin{tabular}{c|ccccc} 
\hline
Method & SqRel$\downarrow$ & rms$\downarrow$ & log10$\downarrow$ & $\delta_{1}\uparrow$ & $\delta_{2}\uparrow$ \\ 
\hline
w/ & 0.468 & 2.977 & 0.051 & 0.833 & 0.952 \\ 
\hline
w/o & 0.791 & 3.811 & 0.062 & 0.799 & 0.952 \\
\hline
\end{tabular}}
\caption{Comparison of our architecture with and without the first convolution layer of high-resolution depth.}\label{tab:e_g_ablation}
%\vspace{-25pt}
\end{table}

%\red{To demonstrate the effectiveness of the 3-layer fusion network, we train a 2-layer version. We adopt LeRes~\cite{yin2021learning} with resNeXt101\cite{xie2017aggregated} as the backbone single-resolution depth prediction method and compare the 2-layer network with the 3-layer network on iBims-1\cite{koch2018evaluation}. The errors are shown in Table~\ref{table:ablation}. We evaluate the absolute mean relative error ($AbsRel$),  root mean square error($rms$) and the boundary metrics $\varepsilon _{DBE}^{acc}$ and $\varepsilon _{DBE}^{comp}$ from \cite{koch2018evaluation}. It shows that the 2-layer version is too simple to complete the fusion well, and compare to the backbone single-resolution method LeRes, the improvement by 2-layer fusion is minor. The 3-layer version in the proposed method boosts the performance under various metrics. }

% \begin{figure}
% 	\centering
% 	\includegraphics[width=0.99\linewidth]{loss.pdf}\
% 	\caption{\red{Visual comparisons of different loss function settings. Our proposed ranking loss can effectively improve the detail(B and D).}}
% 	\label{fig:loss}
% \end{figure}

% \begin{figure}
% 	\centering
% 	\includegraphics[width=0.99\linewidth]{classcompare.pdf}\
% 	\caption{.}
% 	\label{fig:texture_plot}
% \end{figure}
\vspace{-3pt}
\subsection{Limitations}
\begin{figure}[!t]
	\centering
	\includegraphics[width=\linewidth]{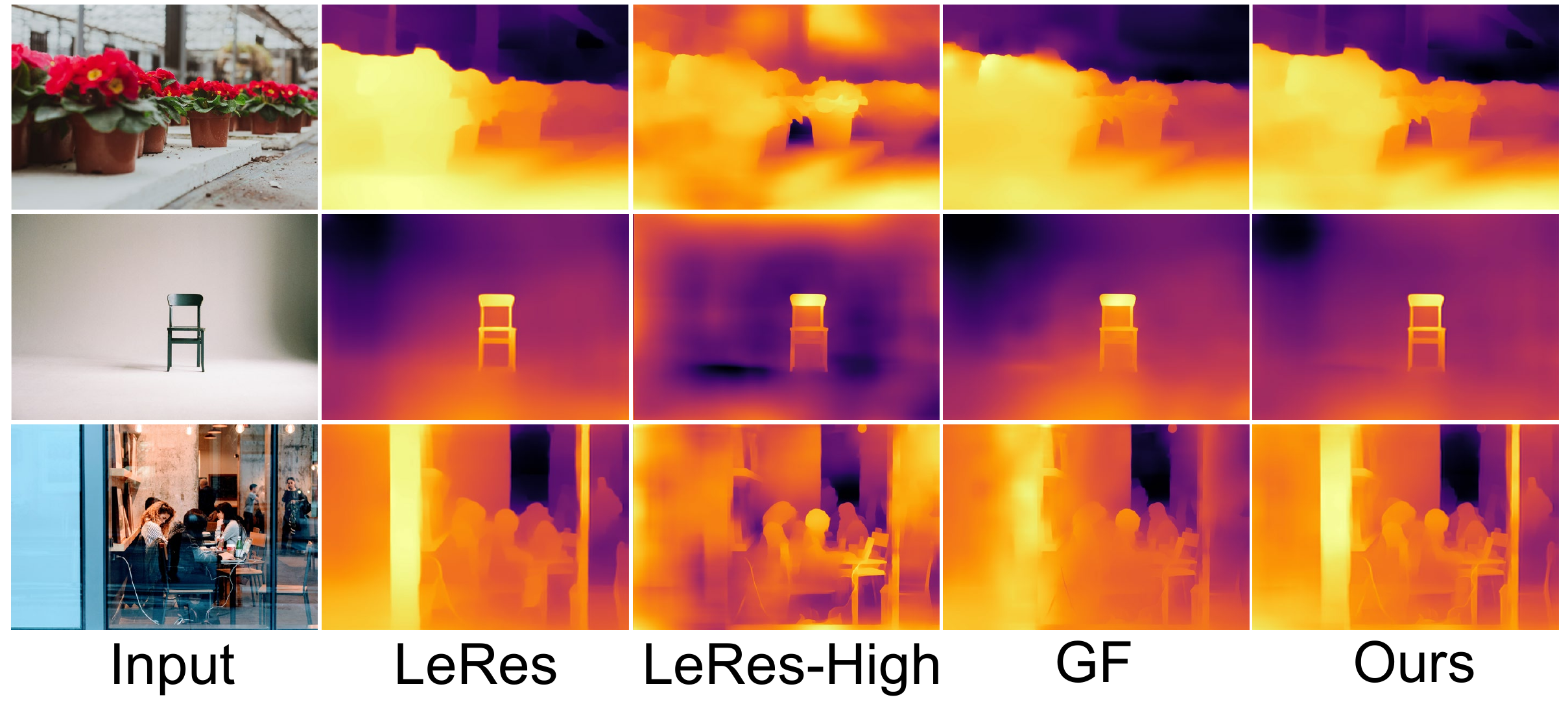}\
	\caption{Three types of challenging cases, which are out-of-focus images, featureless regions, and transparent surfaces. }
	\label{fig:failcase}
	\vspace{-15pt}
\end{figure}
%Since our method is based on any backbone monocular depth estimation and improves their results with much more details. A problem is when the backbone method generates inaccurate depth predictions, in order words, when the three-resolution depth maps are all implausible in depth values. Our method cannot fix the inaccurate depth values and still get a plausible fusion results. 
Our method benefits from details in high-resolution images. 
%Our method fuses details recovered from depth predictions of high-resolution images into predictions of low-resolution images, to improve the performance of the backbone method. However 
If the original resolution of input images are low, the improvements by our multi-resolution depth fusion are minor. Furthermore, there are other challenging cases for our method. 
%We conclude three scenarios that the depth enhancement by our method is not significant. 
As demonstrated in Fig.~\ref{fig:failcase}, the first scenario is out-of-focus images (the $1^{st}$ row). Since the out-of-focus regions are blurred, the details cannot be recovered by high resolution depth map. The second type is the featureless images. As shown in $2^{nd}$ row, the white floor and wall are featureless, and monocular depth estimation backbones cannot predict the accurate depth. The $3^{rd}$ case is transparent or reflective materials such as water and glasses. Since there is a glass wall in front of the scene, the depth enhancement results contain many artifacts.

 %Moreover, with recent successes of introducing scene analysis into 3D vision tasks, our method may be better incorporating with scene analysis such as edges, boundaries or plane detection. Introducing semantics into this task is also a promising future direction. 
%Semantic information that can be predicted to better supervise or constrain the monocular depth estimation problem. 
%Incorporating this problem with another highly-related problem, image segmentation, is also worthy to explore. Both tasks can be very helpful to each other, and a joint training may lead to boosts of both tasks. 
% \begin{table}[t]
% \centering
% \scalebox{0.7}{
% \begin{tabular}{c|ccc|c} 
% \hline
% \multirow{2}{*}{} & \multicolumn{3}{c|}{Loss function} & Result \\ 
% \cline{2-5}
%  & ILNR & Gradient & Ranking & error \\ 
% \hline
% Setting A & Low-res depth & High-res depth & x & 0.734 \\
% Setting B & Low-res depth & \cross & High-res depth & \uline{0.688} \\
% Setting C & Guided depth & Guided depth & x & 0.723 \\
% \hline
% Setting D & Guided depth & x & Guided depth & \textbf{0.684} \\
% \hline

% \end{tabular}}

% \label{table:ablation} 
% \end{table}

	\vspace{-3pt}
	\section{Conclusion}
	
	We introduce a multi-resolution gradient-based depth map fusion pipeline to enhance the depth maps by backbone monocular depth estimation methods. 
	%Our method improves the depth maps by backbone methods both quantitatively and qualitatively. 
	Depth maps with a wide level of details are recovered by our method, which are helpful for many following and highly-related tasks such as image segmentation or 3D scene reconstruction. Comparing with prior works in depth map fusion, the proposed method has a great robustness to image noises, and runs in real time. The self-supervised training scheme also enables training on unlabeled images. Comprehensive evaluations and a large amount of ablations are conducted, proving the effectiveness of every critical modules in the proposed method. %In the future, we plan to further explore high-resolution depth estimation with semantics analysis. 
	\section*{Acknowledgement}
	
	We thank the anonymous reviewers for their valuable comments. This work is supported in part by the National Key Research and Development Program of China (2018AAA0102200), NSFC (62132021, 62002375, 62002376), Natural Science Foundation of Hunan Province of China (2021JJ40696, 2022RC1104) and NUDT Research Grants (ZK19-30, ZK22-52).

	\clearpage
	\begin{appendices}
		
		\section{Supplementary material}
		\subsection{Discussions on the multi-resolution theory}
		
		\begin{figure}[h]
			\centering
			\includegraphics[width=0.8\linewidth]{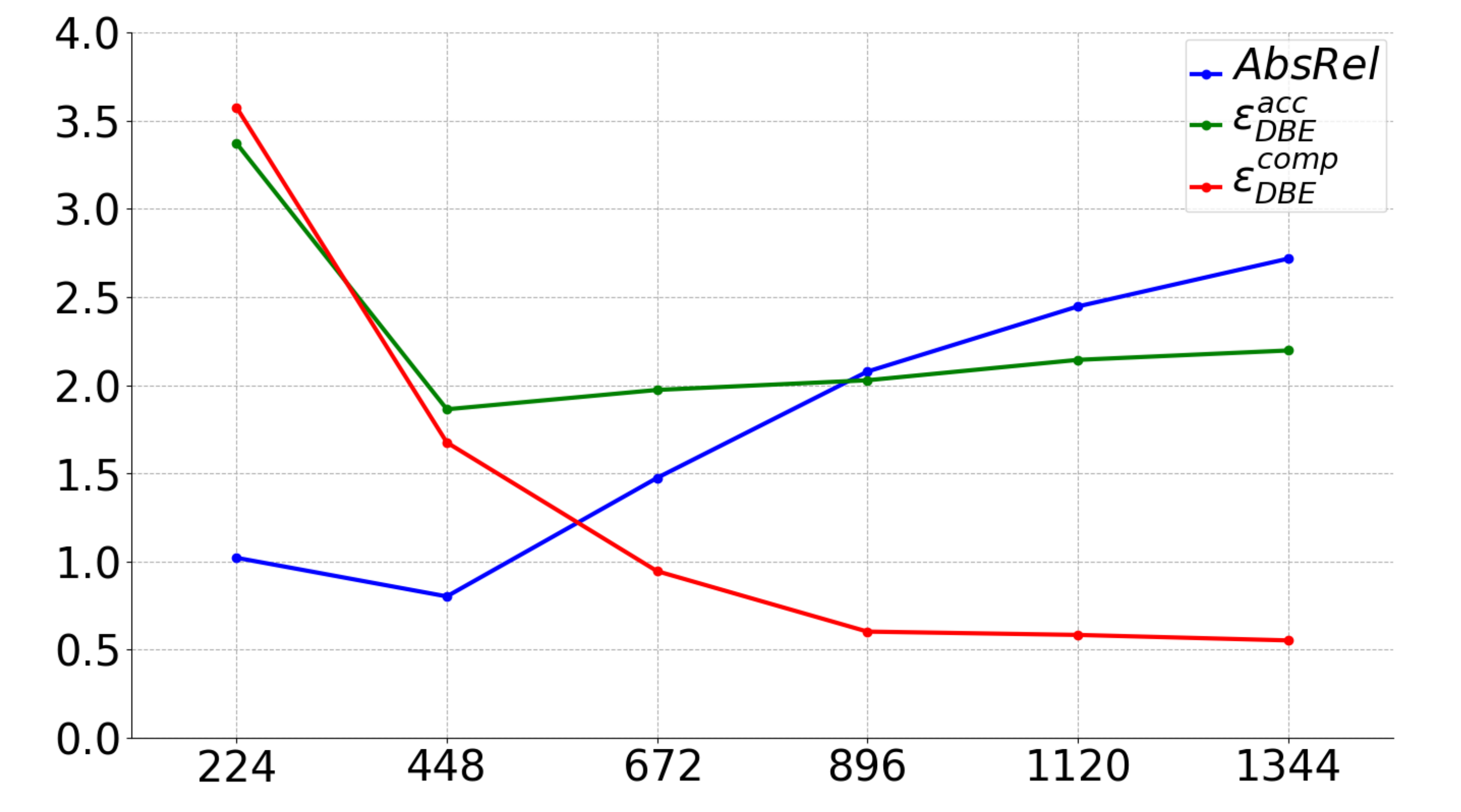}\
			\caption{Evaluations of LeRes predictions on different input resolutions. X-axis denotes input resolutions, and y-axis denotes values of error metrics after scaling into the same range.}
			\label{fig:error}
		\end{figure}
\begin{figure*}[h]
			\centering
			\includegraphics[width=0.99\linewidth]{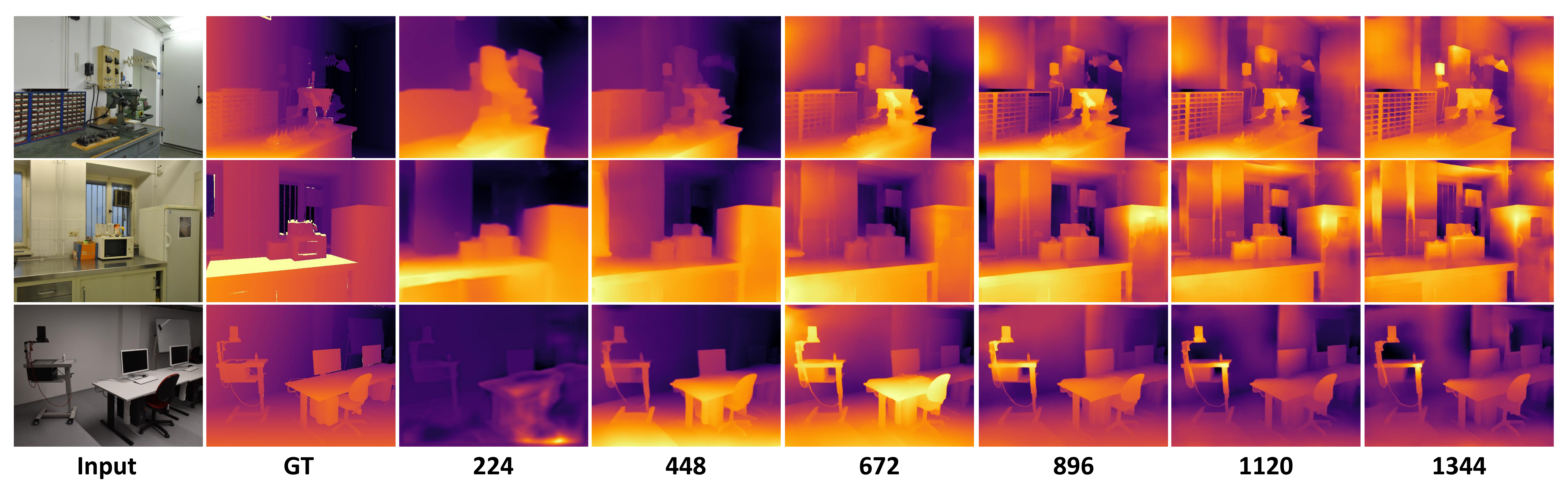}\
			\caption{Visualizations of depth estimations at different input resolutions by LeRes~\cite{yin2021learning}.}
			\label{fig:res}
		\end{figure*}

 
		Our work is based on the observation that for fully convolutional monocular depth estimation networks, input images at training resolutions produce the highest accuracy in depth values, while high-resolution inputs lead to more details with inaccurate depth values. In order to test this multi-resolution theory, we adopt input images at 6 different resolutions to predict corresponding depth maps at different resolutions. We evaluate these predicted depth maps by several metrics. The evaluation is conducted on Ibims-1~\cite{koch2018evaluation} dataset, which provides images at the resolution of $640\times480$. We utilize down-sampling and linear interpolation to change the resolutions from $224\times224$ to $1344\times1344$. We evaluate three metrics $AbsRel$, $\mathcal{E}^{acc}_{DBE}$ and  $\varepsilon^{comp}_{DBE}$ provided by Ibims-1. While $AbsRel$ evaluate the global accuracy of the depth map, $\varepsilon^{acc}_{DBE}$ and  $\mathcal{E}^{comp}_{DBE}$ evaluate the depth boundary accuracy error and the depth boundary completeness error, respectively. The three metrics are the lower the better. We test the depth prediction of LeRes at 6 different resolutions, the error values are plotted in Figure~\ref{fig:error} with qualitative results demonstrated in Figure~\ref{fig:res}. The training resolution of LeRes is $448\times448$, this resolution at Figure~\ref{fig:error} achieves the lowest $AbsRel$ proving the correctness of our method to fuse depth values from training resolutions (low-resolution in our method). Inputs smaller than $448\times448$ are incorrect in both accuracy and details (high in all three metrics), and inputs larger than $448\times448$ lead to good details (low in  $\varepsilon^{comp}_{DBE}$) with inaccurate depth values (high in $AbsRel$).  Since some boundaries are missing in the ground truths depth map, the boundary accuracy $\varepsilon_{DBE}^{acc}$~\cite{koch2018evaluation} increases a little bit. %This is why multi-resolution depth estimations should be fused to get an overall plausible one. 
		It is a common observation of fully convolutional neural networks. As long as two depth maps at training resolution and higher resolution are given, our fusion network can successfully fuse them into an overall plausible one by keeping both value accuracy and details, without any finetuning on backbone networks. It is why our depth fusion pipeline has a great generalization ability while changing backbones, no additional finetuning is needed. 
		
		\subsection{Technical and implementation details}

		\subsubsection{Implementation details}\label{sec:implementation}
		While incorporating on the backbone of LeRes~\cite{yin2021learning}, here we provide more details. Bilinear interpolation is adopted to generate the high-resolution input for our fusion framework from the original input. The training loss $l_{\text{mILNR}}$ and $l_{\text{rank}}$ is constructed with three levels of resolution, which are 512, 1024 and 2048. 
		
		The detailed network structure of our fusion network is provided in Figure~\ref{fig:struc}. 
		Our network is end-to-end optimized using the AdamW \cite{loshchilov2017decoupled} optimizer with the batch size of 2, using HR-WSI~\cite{xian2020structure} dataset as the training set. The initial learning rate is 0.0001, and 2 epochs of training on the whole 36 training sets are processed. For ablation study we trained on one training set for 30 epochs. The Cosine Annealing \cite{loshchilov2016sgdr} is adopted as the learning rate scheduler, and the decay rate is fixed as 0.99 every 100 steps. We implement our method on the Ubuntu22 system with one NVIDIA TITAN V GPU and a 16G RAM. The CUDA version is 11.1 and the Python version is 3.8. We use the depth prediction architecture proposed in LeRes~\cite{yin2021learning} as monocular depth estimation backbone, consisting of two versions of encoders (ResNet50~\cite{he2016deep} or ResNeXt101~\cite{xie2017aggregated}) and a decoder. We adopt ResNet50 as our encoder backbone through the whole evaluation process. To preprocess the HR-WSI dataset, the low-resolution inputs are resized to $448\times448$, which is the default resolution of LeRes. High-resolution inputs are resized to $1344\times1344$ to predict missing details. For radius and regularization parameters in guided filter, we fixed them as $\frac{W}{12}$ and $10^{-12}$ ($W$ denote the width of the input image). We find that radius larger than this value would cause blurry and regularization larger than this value would lose details. We use canny edge detection to select 7200 high quality guided filter results as our training set. The regular term $\sigma$ of ranking loss is fixed as 0.1, after testing values between $[0,1]$. 
		%, whose value range from 0 to 1 to get a smooth loss function.} 
		%The parameters for generating the edge mask $M$ for $l_{\text{rank}}$ are set default the same as \cite{loshchilov2016sgdr}. 
		%During our self-supervised training, we freeze parameters of the backbone throughout the whole training period to train the fusion network. 
		
		%While incorporating with other two different backbones,  SGR~\cite{xian2020structure} and MiDas~\cite{ranftl2019towards}.\yaqiao{(delete this sentence)} 
		While LeRes and NeWCRFs predict inverse depth values, SGR, MiDas and DPT directly predict depth values. Thus, while using our fusion network trained with LeRes backbone, We transform the depth values predicted by SGR, MiDas and DPT to inverse depth by $D_{inverse} = D_{max}-D$ ($D_{max}$ is the maximum depth value in the whole depth map) before sending to the fusion network. The low resolution depth predictions of SGR,  MiDas, DPT and NeWCRFs are set to $448\times448$,  $392\times392$, $384\times384$ and $640\times480$  respectively, which are the same as their training resolution. For high resolution depth predictions, we set the resolution as $1344\times1344$, $1176\times1176$, $1152\times1152$ and  $1920\times1440$ respectively. Changing the backbone from LeRes to others does not require any additional finetuning.  All experimental results are got by testing for one time. 
		
		\subsubsection{Details of training strategies}
		For robust training, we apply several training strategy to improve generalize ability of the model. Before fusion depth, we first scale the depth value by $D_{scaled} = MinMax(D_{pred})\times2 - 1$ ($MinMax(D) = (D-D_{min})/(D_{max} - D_{min})$).We adopt data augmentation of random flipping of x,y axes and random transforming to inverse depth by $D_{input} = -D_{scaled}$ ($D_{scaled}$ is the depth value scaled to range $[-1,1]$.). Although the fusion network is trained to fuse both depth maps and inverse depth maps, while testing we feed inverse depth maps for all backbones for a fair comparison.

		Besides, we adopt a multi-scale training strategy to supervise the intermediate results of the proposed fusion network. We adopt the same structure of our last decoding block but with different input channel number, to predict two additional fused depth maps at lower resolutions ($1024\times1024$ and $512\times512$). In our training process, the three fused depth maps at different resolutions are used to calculate the losses $l_{\text{mILNR}}$ and $l_{\text{rank}}$. 
		
		\begin{figure*}
			\centering
			\includegraphics[width=0.99\linewidth]{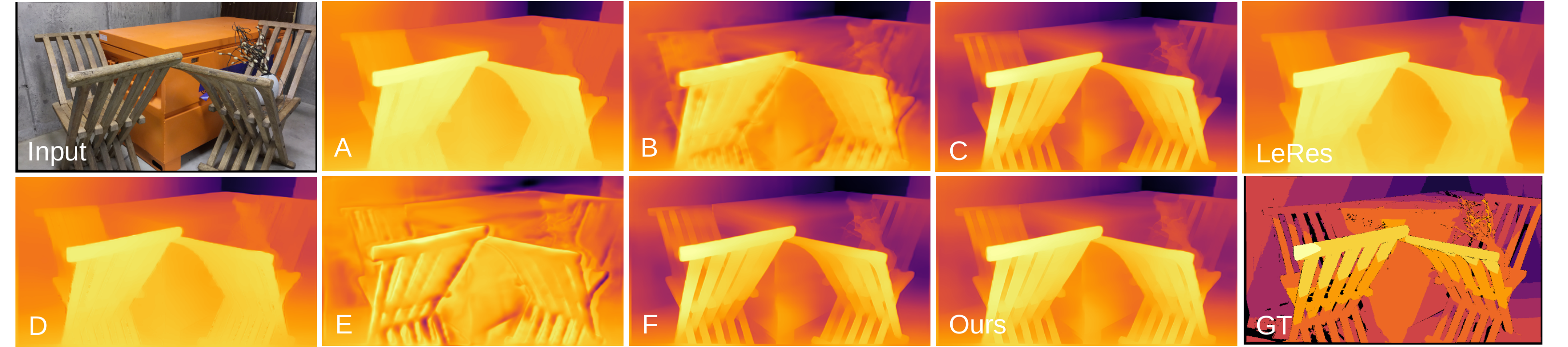}\
			\caption{Visual comparison of different loss function settings. Our loss function design recovers the details correctly, while using ILNR loss (C and F) to learn the guided filter result leads to incorrect artifacts near the wall and desk.}
			\label{fig:loss}
		\end{figure*}
		
		\begin{figure}
			\centering
			\includegraphics[width=1\linewidth]{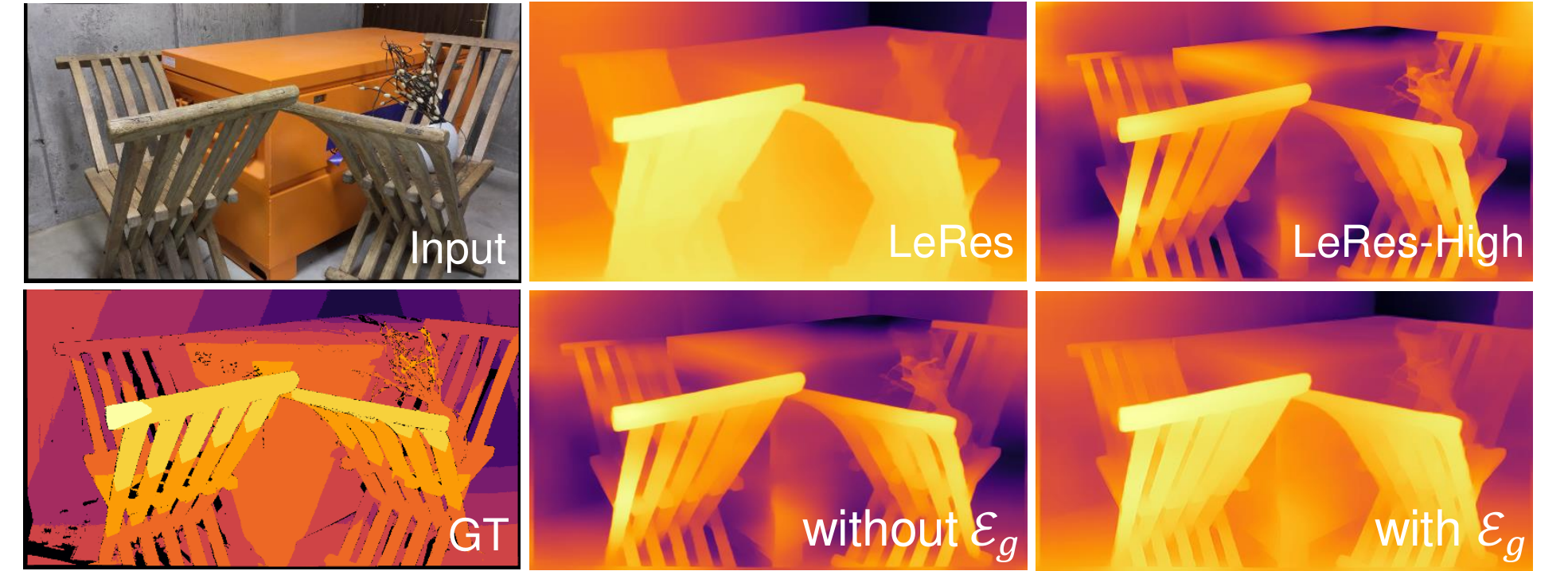}\
			\caption{Visual comparisons of results with and without $\mathcal{E}_g$. Result without $\mathcal{E}_g$ contains many wrong depth relations from high resolution depth.}
			\label{fig:noconvcompare_supp}
		\end{figure}
		
		\subsubsection{Sampling strategy of our ranking loss}
		Ranking loss is a widely used loss function for many metric learning tasks. It is constructed based on sampled pairs of pixels. However, previous ranking loss only focus on keeping the order while the value error is essential in our method for supervising the training of the gradient domain. We introduce a novel ranking loss which can constrain the value error as well as the order. Sample strategy of our method are presented below. 
		
		For monocular depth estimation task, DIW~\cite{chen2016single} adopt the pair-wise ranking loss to improve the prediction accuracy with a random sample strategy. SGR~\cite{xian2020structure} propose the structure guided ranking loss by sampling points around instance boundaries. We follow the edge-guided sampling strategy of SGR~\cite{xian2020structure}, sample point-pairs near the depth boundaries. The depth boundaries are generated follow the pipeline of canny edge detection without denoise, and change the non-maximum suppression to  threshold. We firstly utilize the Sobel operator to calculate the gradient maps of a depth map generated by guided filter we mentioned in our paper. After we get the gradient maps $G_x$, $G_y$ and gradient magnitude map $G$, we compute the depth boundaries map $E_d$ by thresholding the gradient magnitude map use the function as follow: $E_d = \mathbb{I}[G\geq(1-\alpha)\cdot \max(G)]$. 
		
		Here $\alpha$ is a threshold parameter to control the density of depth boundaries map. For each point $e=(x, y)$ in $E_d$, we sample four points $[e_k = (x_k, y_k), k=a, b, c, d] $ alongside the gradient direction at point $e$ by equations as follow:
		
		\begin{equation}
			\left\{
			\begin{array}{lr}
				x_k= x + \delta_k G_x(e)/G(e)  \\
				y_k= y + \delta_k G_y(e)/G(e)
			\end{array}
			\right.
		\end{equation}
		
		The parameters $\delta_k$ are generated randomly while make sure the sampled points are near the depth boundaries point $e$, and they are ordered as $\delta_a \textless \delta_b \textless 0 \textless \delta_c \textless \delta_d $. To avoid sampled too far, we limit that $\left \| \delta_k\right \| \leq \beta$. We finally add $(e_a, e_b), (e_b, e_c) and (e_c, e_d)$ into the point-pair set to calculate our novel ranking loss. We also apply the same sample strategy to $D_{high}$ and combine the point-pair set together. The calculation weights of point-pair set from $D_{gf}$ and $D_{high}$ are set as 12 and 8.
		
		We use a indicator $Z_{ij}$ to define the relation of the point-pair $(e_i, e_j)$ which is 
		\begin{equation}
			\left\{
			\begin{array}{ll}
				z_{ij}= 1, &\frac{p_{gf}^i}{p_{gf}^j} \geq 1 + \tau \; or \; \frac{p_{gf}^i}{p_{gf}^j} \leq  \frac{1}{1 + \tau} \\
				z_{ij}= 0, & otherwise
			\end{array}
			\right.
		\end{equation}

\begin{figure*}
			\centering
			\includegraphics[width=0.8\linewidth]{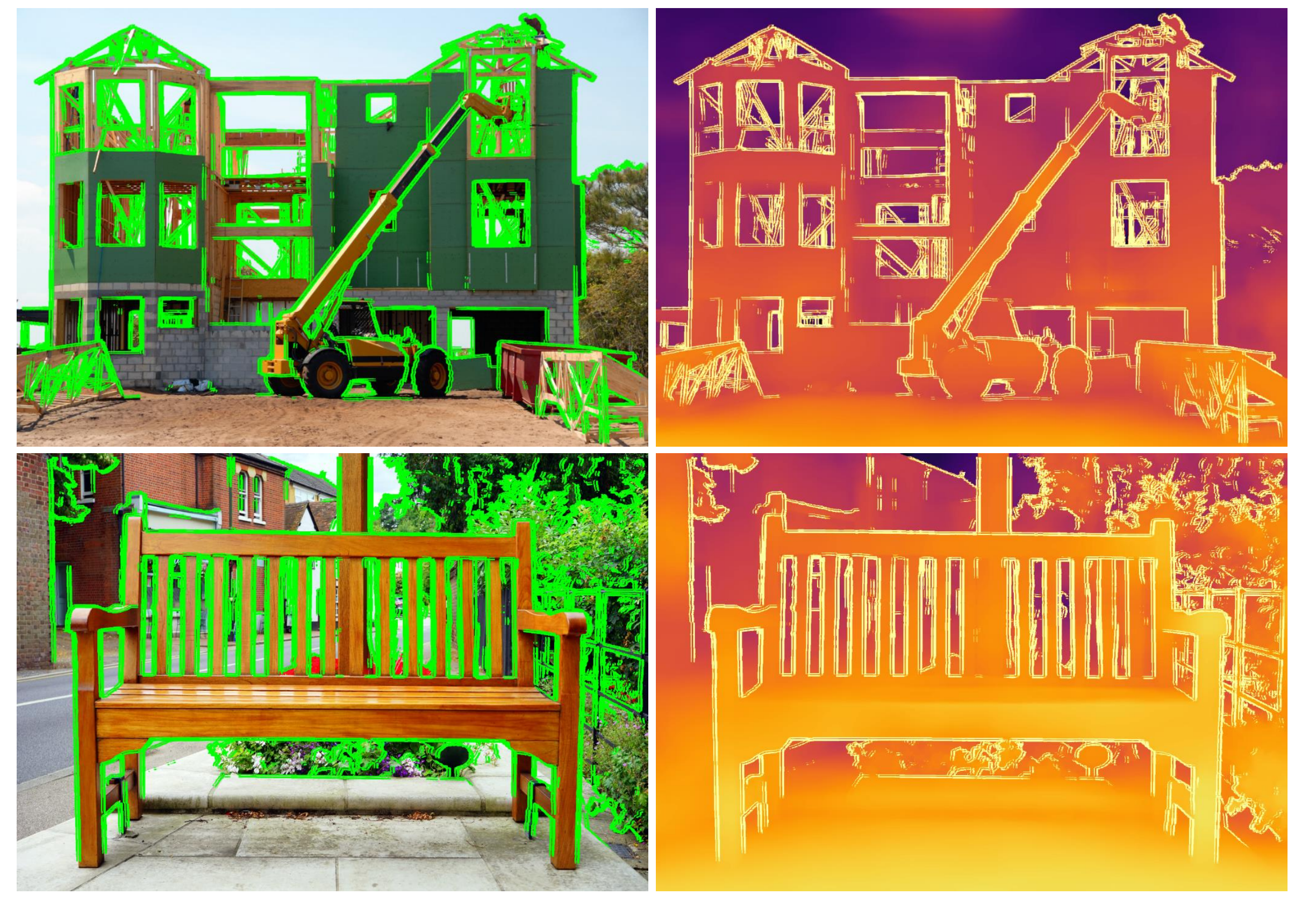}\
			\caption{Visualizations of sample points. Here we set $\beta$ to 10 which makes the sample points being close to edges where gradients change most rapidly for better illustration. }
			\label{fig:samplePoint}
		\end{figure*}

		\noindent $\tau$ is the tolerance threshold of depth boundary which we set to 0.001. In our training process, we set $\alpha$ and $\beta$ as 0.15 and 60 respectively. For lower fused depth ($1024\times1024$ and $512\times512$) we set $\beta$ as 30 and 15. Other parameters are introduced in Section 3.3 of the main paper. %\yaqiao{The meaning of other characters are mentioned in the section 3.3 of our main paper.} 
		Figure~\ref{fig:samplePoint} gives out some visualization example of sampled points, which in input image are colored as green (left) while in depth map are colored as beige (right).

		\subsection{Detailed information of benchmarks}
		Here we introduce several widely-used image datasets with depth ground truths.
		
		Multiscopic~\cite{yuan2021stereo} dataset is an indoor dataset including real data collected by multi-lens cameras and synthetic data generated by renderers. The dataset contains 1100 scenes of synthetic data for training, 100 scenes of synthetic data for testing and 100 scenes of real data for other tasks. For each synthetic scene, Multiscopic~\cite{yuan2021stereo} provide 5 different views of RGB image and disparities, at a resolution of $1280\times1080$. For each real scene, 3 different views of RGB images and disparities are provided.
		
		Middlebury2021~\cite{scharstein2014high} is a high quality stereo dataset collected by a structured light acquisition system. This dataset provides images and disparities for different indoor scenes at a resolution of $1920\times1080$. The dataset contains 24 testing scenes with their calibration files respectively. At test time, we transform the disparity value $d$ to depth value $z$ by $z = baseline * f / (d + d_{offs})$, where the values of $baseline$, the focal length $f$ and the displacement of principal points $d_{offs}$ are provided in the dataset.
		
		Hypersim~\cite{roberts:2021} is a photorealistic synthetic dataset for holistic indoor scene understanding. It contains 77400 images of 461 indoor scenes with per-pixel labels and corresponding ground truth geometry. Due to the large size of the whole dataset, we test on three subsets containing 286 tone-mapped images generated from their released codes.
		
		NYUv2~\cite{Silberman:ECCV12} dataset is a wildly-used indoor dataset that provides real data at the resolution of $640\times480$. Data in the NYUv2 are captured by RGB and Depth cameras in the Microsoft Kinect. It includes 120K training images and 654 testing images. In evaluation, we use their official test split.
		
		KITTI~\cite{Uhrig2017THREEDV} is a outdoor stereo depth map dataset,  providing thousands of raw LiDar scans and their corresponding RGB images. The stereo images and 3D laser scans of street scenes are captured by a moving vehicle. Images are at the resolution of $1216\times352$. While the aspect ratio of these data is about 4:1, the depth maps collected by laser equipment exist a large amount of missing points. We test on its testing subset of 652 images.
		
		For evaluation metrics mentioned in the main paper, the square relative error $SqRel$ is defined as $SqRel=\frac{1}{N}\sum(\frac{d_{i}^{*}-d_{i}}{d_{i}^{*}})^{2}$. 
		The root mean square error $rms$ is defined as 
		$rms=\sqrt{\frac{1}{N}\sum {\left \|d_{i}^{*}-d_{i}\right \|}^{2}}$. 
		
		\subsection{Additional experiments and results}
		
		\subsubsection{Evaluation on the backbone of Midas}
		In the paper, we provide the quantitative evaluation of our method on four monocular backbones, comparing with other fusion methods. Here We also provide the quantitative evaluation of our method on the backbone of MiDas to show the performance boost by our method. Quantitative results are in the Table ~\ref{tab:midas}. Besides the evaluation metrics mentioned in the paper, we also evaluate our method on MiDas using $ORD$~\cite{xian2020structure} and $D^{3}R$~\cite{miangoleh2021boosting}, which using the same evaluation function but with different sample strategy. The evaluation function of $D^{3}R$ is as follow:
		$D^{3}R=\frac{\sum_{k}w_{k}\mathbb{I}(Z_{ij} \neq Z^{*}_{ij})}{\sum_{k}w_{k}} $ , where $w_k$ is a weight set to 1 and $Z_{ij}$ denote the sampled pair-points relation of $e_i$ and $e_j$. 
		\begin{table}[h]
			\centering
			\scalebox{0.7}{
				\begin{tabular}{c|ccccccc} 
					\hline
					\multirow{2}{*}{Methods} & \multicolumn{7}{c}{Multiscopic}                                                                                                                                             \\ 
					\cline{2-8}
					& SqRel$\downarrow$& rms$\downarrow$& log10$\downarrow$& $\delta_1\uparrow$& $\delta_2\uparrow$          & ORD$\downarrow$           & $D^3R \downarrow$        \\ 
					\hline
					MiDas                    & 6.118          & 11.684 & \textbf{0.070} & 0.800 & 0.924 & 0.192          & 0.557                    \\ 
					\hline
					MiDas-GF                    & 6.204          & 11.694          & \textbf{0.070} & 0.800 & 0.924 & 0.194          & 0.549                    \\ 
					\hline
					MiDas-Ours& \textbf{6.034} & \textbf{11.602}          & \textbf{0.070}          & \textbf{0.801} & \textbf{0.925} & \textbf{0.190} & \textbf{0.536}   \\
					
					% \end{tabular}}
					% \caption{Evaluation of our fusion module with MiDas on Multiscopic and Middlebury2021. }
					% \label{tab:midas}
					% % \vspace{-325pt}
					% \end{table}
					\hline 
					
					\multirow{2}{*}{Methods} & \multicolumn{7}{c}{Middlebury2021}\\ 
					\cline{2-8}
					& SqRel$\downarrow$ & rms$\downarrow$ & log10$\downarrow$ & $\delta_1\uparrow$ & $\delta_2\uparrow$ & ORD$\downarrow$ & $D^3R\downarrow$  \\ 
					\hline
					MiDas & 0.750 & 3.766 & 0.063 & 0.811 & 0.947 & 0.249 & 0.713 \\ 
					\hline
					MiDas-GF & 0.737 & 3.722 & \textbf{0.062} & 0.810 & 0.948 & 0.247 & 0.712 \\ 
					\hline
					MiDas-Ours & \textbf{0.72} & \textbf{3.67} & \textbf{0.062} & \textbf{0.816} & \textbf{0.949} & \textbf{0.244} & \textbf{0.703}\\
					\hline
					
					% \begin{table*}[h]
					% \centering
					% \scalebox{1}{
					% \begin{tabular}{c|ccccccc|ccccccc} 
					
					\multirow{2}{*}{Methods} & \multicolumn{7}{c}{Hypersim}\\ 
					\cline{2-8}
					& SqRel$\downarrow$ & rms$\downarrow$ & log10$\downarrow$ & $\delta_1\uparrow$ & $\delta_2\uparrow$ & ORD$\downarrow$ & $D^3R\downarrow$  \\ 
					\hline
					MiDas & 0.445 & 1.298 & 0.091 & 0.69 & 0.893 & 0.216 & 0.452  \\ 
					\hline
					MiDas-GF & 0.444 & 1.298 & 0.092 & 0.689 & 0.893 & 0.219 & 0.424 \\ 
					\hline
					MiDas-Ours & \textbf{0.428} & \textbf{1.275} & \textbf{0.09} & \textbf{0.694} & \textbf{0.895} & \textbf{0.212} & \textbf{0.396}\\
					\hline
			\end{tabular}}
			\caption{The quantitative evaluations on three benchmark datasets, comparing to a monocular depth estimation network MiDas~\cite{ranftl2019towards}, guided filter (``GF'' in the table) and our method on the backbone of MiDas. Bold numbers denote the best performance of each metric. Our method achieve the best performance in these three benchmarks.}
			\label{tab:midas}
			% \vspace{-325pt}
		\end{table}

		\subsubsection{Visual comparisons}
		Visualizations of depth enhancements on three backbones are in Figure~\ref{fig:visuMidd} (for Middburry dataset) and Figure~\ref{fig:visuHyper} (for Hypersim dataset). We also provide depth fusion results of our methods on three backbones on many Internet images in Figure~\ref{fig:visuWild_1}-\ref{fig:visuWild_2}. 
		%Since Internet images are at high resolutions, we define the high resolution as $2240\times2240$ ($1920\times1920$ for MiDas) for high resolution depth estimation, to keep the most details. 
		We can see from the figures that since images in Middburry dataset are at the resolution of $1920\times1080$, the visual improvements by depth fusion are more obvious than low resolution images of $1024\times768$ in Hypersim dataset. From the Internet images, the improvements are even better, since Internet images are at higher resolutions from $2590\times1680$ to $8192\times5462$. Therefore, our method produces more depth improvements on high resolution inputs, which is also the target scenario of the proposed method.   %Since images from different datasets are various in sizes, we define the high resolution of image as $1344\times1344$ ($1176\times1176$ for MiDas) for images whose original size is smaller than $1920\times1080$. 

		We also test our method on several Internet videos (see the videos shown on {\color{green}{\url{https://github.com/yuinsky/gradient-based-depth-map-fusion}}}), by performing the proposed single-image pipeline frame by frame. Since each frame is processed separately, there are some inconsistencies among frames. 
		
		As introduced in ablation studies in the main paper, We also visualize results of different loss function settings (details about loss settings A to F are mentioned in Section 4.3 in the main paper) in Figure ~\ref{fig:loss}. We can see from the visualization results that using $l_{\text{mILNR}}$ to learn guided filter result leads to incorrect artifacts. We also visualize the depth fusion results with and without $\mathcal{E}_g$ in Figure~\ref{fig:noconvcompare_supp}, from which we found that the fused depth would not follow the depth values of low resolution depth while not using $\mathcal{E}_g$.

		\subsubsection{Noise robustness}
		
		As described in Section 4.2 of the main paper, we evaluate the anti-noise ability by testing on images with synthetic Gaussian or Pepper noises. 
		Gaussian noises often exist when the field of view of the image sensor is not bright enough or the brightness is not uniform enough when shooting. The image sensor works for a long time and the temperature of the sensor is too high may also cause such noises. Salt and pepper noises, also known as impulse noises, are often caused by image clipping, randomly changes some pixel values to black or white. Such spot noise may also be generated by strong interference on the signal of image sensors and errors occurred in transmission or decoding process, etc. 
		
\begin{figure*}
			\centering
			\includegraphics[width=0.99\linewidth]{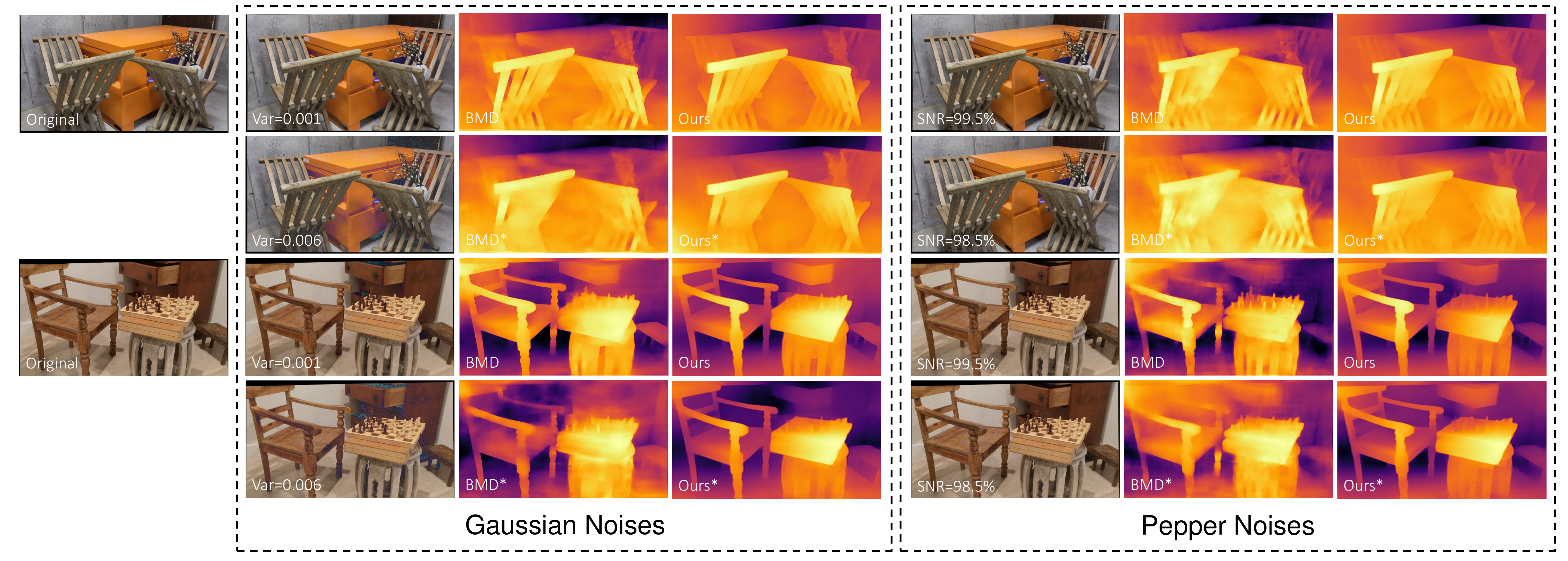}\
			%\vspace{-15pt}
			\caption{Illustrations of our results with different levels of Gaussian or Pepper noises, compared with a depth fusion alternative BMD~\cite{miangoleh2021boosting}. }
			\label{fig:pepper}
		\end{figure*}

\begin{figure*}[h]
			\centering
			\includegraphics[width=0.4\linewidth]{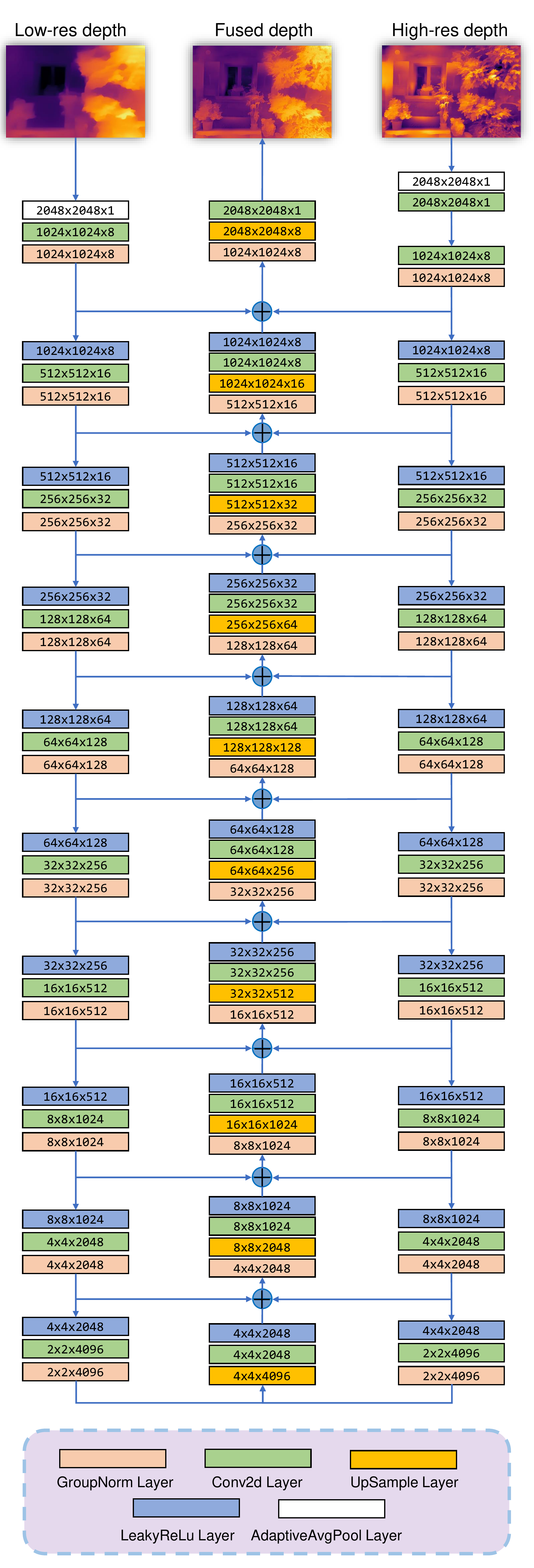}\
			\caption{The network structure of our gradient-based composition module (the fusion network).}
			\label{fig:struc}
		\end{figure*}

\begin{figure*}[h]
			\centering
			\includegraphics[width=0.99\linewidth]{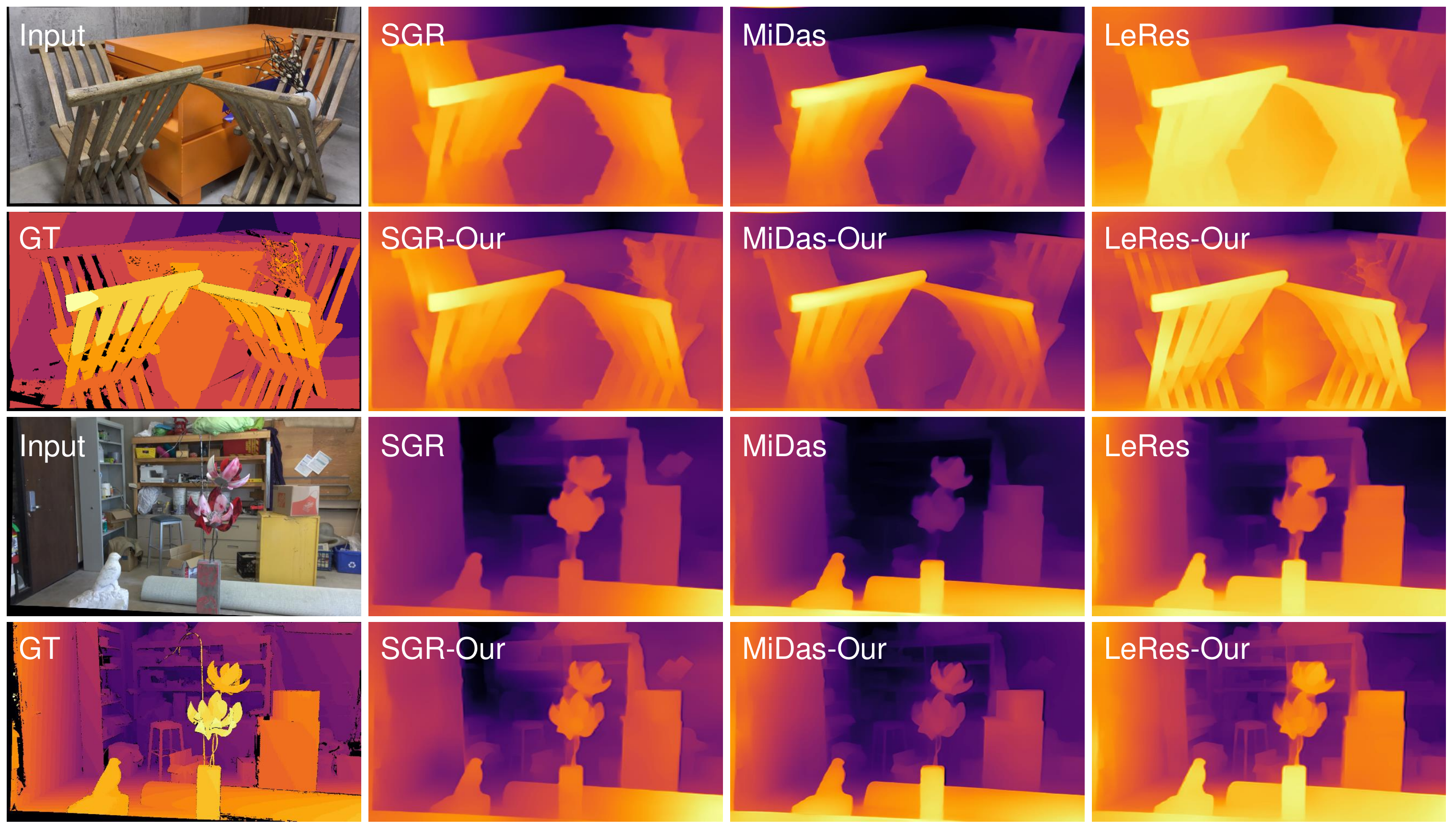}\
			\caption{Visual results of data from Middlebury2021 dataset. 
			}
			\label{fig:visuMidd}
		\end{figure*}

\begin{figure*}[h]
			\centering
			\includegraphics[width=0.99\linewidth]{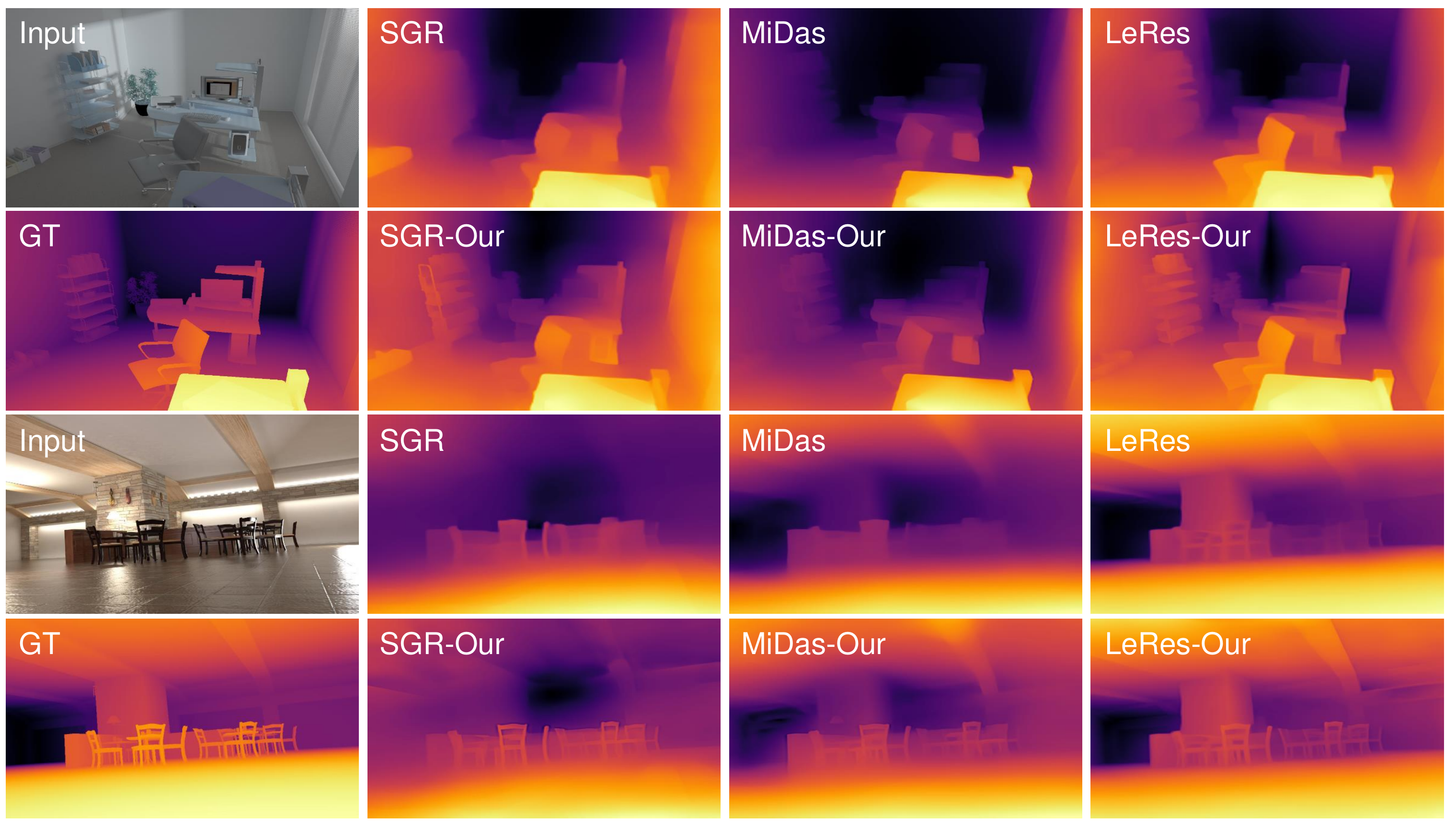}\
			\caption{Visual results of data from Hypersim dataset. 
			}
			\label{fig:visuHyper}
		\end{figure*}

\begin{figure*}[h]
			\centering
			\includegraphics[width=0.99\linewidth]{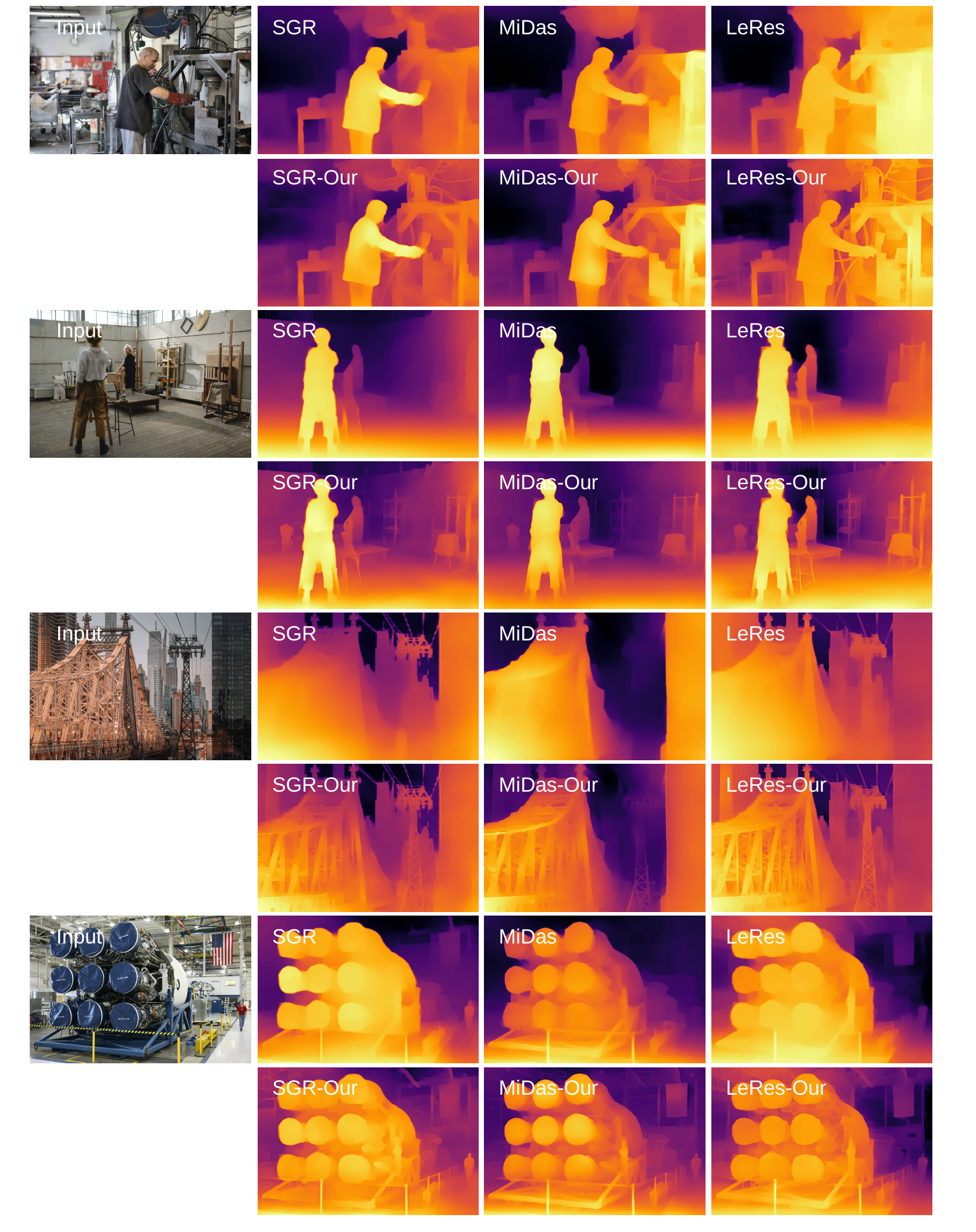}\
			\caption{Visual results on Internet images.}
			\label{fig:visuWild_1}
		\end{figure*}

\begin{figure*}[h]
			\centering
			\includegraphics[width=0.99\linewidth]{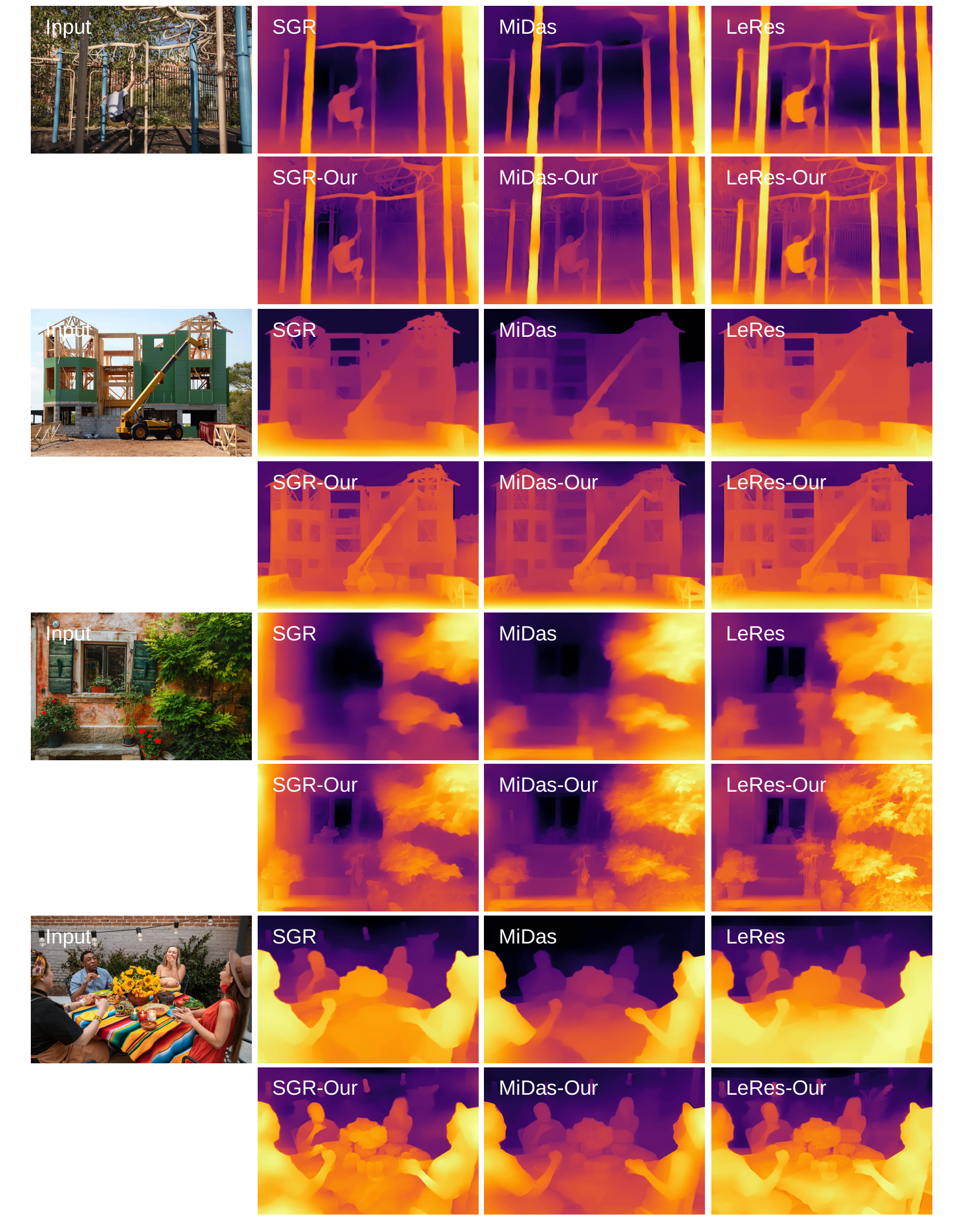}\
			\caption{Visual results on Internet images.}
			\label{fig:visuWild_3}
		\end{figure*}

\begin{figure*}[h]
			\centering
			\includegraphics[width=0.99\linewidth]{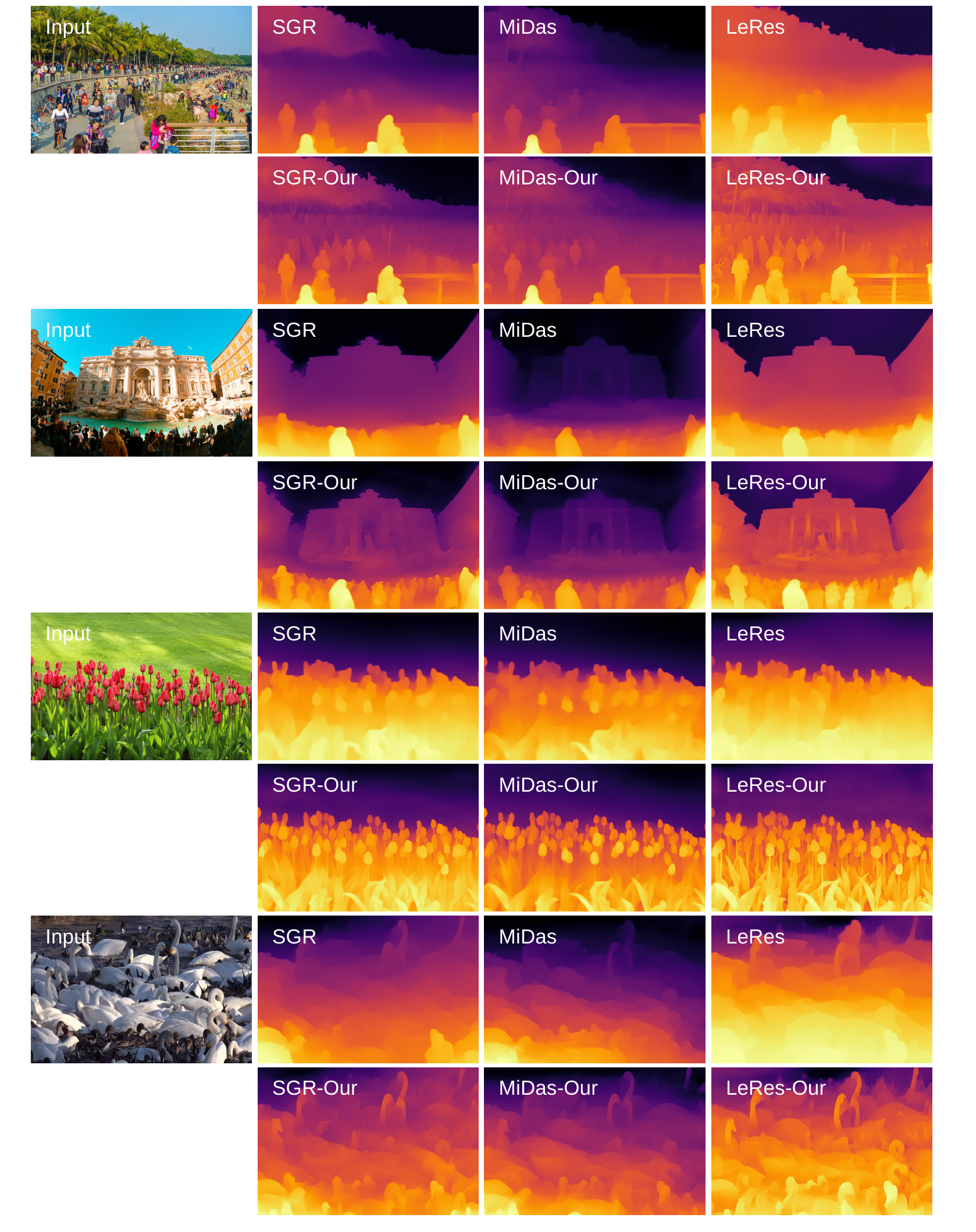}\
			\caption{Visual results on Internet images.}
			\label{fig:visuWild_2}
		\end{figure*}

		Other than the error curves in Figure 8 of the main paper, here we provide two visual examples for better illustration. As in Figure~\ref{fig:pepper}, BMD~\cite{miangoleh2021boosting} suffers from image noises, while our method produce on-par results with and without noises. 
		
	\begin{table}
		\centering
		\label{tab}
		\scalebox{0.85}{
			\begin{tabular}{c|ccccc} 
				\hline
				\multirow{2}{*}{Methods} & \multicolumn{5}{c}{NYU}\\ 
				\cline{2-6}
				& SqRel$\downarrow$ & rms$\downarrow$ & log10$\downarrow$ & $\delta_{1}\uparrow$ & $\delta_{2}\uparrow$ \\
				\hline
				LeRes & 0.134 & \textbf{0.067} & \textbf{0.085} & \textbf{0.783} & \textbf{0.899}\\
				LeRes-GF & \uline{0.128} & \uline{0.068} & \uline{0.087} & \uline{0.7756} & \uline{0.896}\\
				LeRes-BMD & 0.148 & 0.083 & 0.108 & 0.701 & 0.864\\
				LeRes-Ours & \textbf{0.122} & 0.073 & 0.094 & 0.748 & 0.885\\
				\hline
				\multirow{2}{*}{Methods} &\multicolumn{5}{c}{KITTI} \\ \cline{2-6}
				& SqRel$\downarrow$ & rms$\downarrow$ & log10$\downarrow$ & $\delta_{1}\uparrow$ & $\delta_{2}\uparrow$ \\
				\hline
				LeRes & \textbf{0.042} & \textbf{0.094} & \textbf{0.143} & \textbf{0.462} & \textbf{0.718} \\
				LeRes-GF & \uline{0.043} & \uline{0.095} & \uline{0.145} & \uline{0.454} & \uline{0.715} \\
				LeRes-BMD & 0.045 & 0.098 & 0.149 & 0.450 & 0.704 \\
				LeRes-Ours & 0.045 & 0.096 & 0.154 & 0.465 & 0.691 \\
				\hline
		\end{tabular}}
		\caption{Quantitative results on two low-resolution datasets NYU and KITTI.}
	\end{table}
		\begin{comment}
		content...
		
		\begin{figure}
		\centering
		\includegraphics[width=0.99\linewidth]{Gnoise.pdf}\
		%\vspace{-15pt}
		\caption{\red{gauss noise example}}
		\label{fig:gauss}
		\end{figure}
		\end{comment}

		\begin{table}[h]
			\scalebox{0.75}{
				\centering
				%\vspace{-15pt}
				\begin{tabular}{l|l|l|l}
					\hline
					Method & Task & Time per frame & GFLOPs \\ 
					\hline
					\multirow{2}{*}{LeRes} & pred low-res depth & 0.0240s & 95.88092G \\ 
					\cline{2-4}
					& pred high-res depth & 0.0958s & 862.92825G \\ 
					\hline
					\multirow{2}{*}{SGR} & pred low-res depth & 0.0282s & 95.88092G \\ 
					\cline{2-4}
					& pred high-res depth & 0.0981s & 862.92825G \\ 
					\hline
					3DK & pred refinement depth & 0.0141s & 118.02850G \\ 
					\hline
					BMD & pred refinement depth & 14.614s & 14834.4784G \\ 
					\hline
					Guided Filter & fuse low-res and high-res & 0.7025s & - \\ 
					\hline
					Ours & fuse low-res and high-res & 0.0653s & 17.68863G\\\hline
			\end{tabular}}
			\caption{Running time statistics. GFLOPs denotes G floating point operations, for evaluating the computational efficiency. }\label{tab:ruuningtime}
			%\vspace{-25pt}
		\end{table}
		
		\subsubsection{Evaluations on KITTI and NYUv2}
		Several depth fusion method including guided filter, BMD and ours are also tested on NYU and KITTI. %Before discuss the results, there are several extra knowledges to clarify. 
		%The operation of down-sampling, change the high-resolution image to a lower resolution, will led to the reduction of the information contained by that image. As an opposite, interpolate a low-resolution image to a higher resolution cannot enrich the image information. 
		
		State-of-the-arts in monocular depth estimation such as LeRes and SGR, will downsample the high-resolution images to a lower training resolution, which for LeRes is $448\times448$. Based on the observation that some information at original resolution would be lost in down-sampling, depth fusion method including guided filter, BMD and ours utilize such lost information to recover a detailed depth map from depth predictions at high resolutions. It means the original resolution of images should not be too low. If the original resolution in dataset is low, we cannot interpolate them to get more details. 
		
		The low-resolution input size of our model is $448\times448$, and  high-resolution input size is set as $1344\times1344$. However, the original resolution of NYU is $640\times480$, which is almost the same as our low-resolution resolution. The original resolution of KITTI is $1216\times352$, which is also two low in height. 
		
		When tested on these datasets, the high-resolution input is obtained by interpolation operation, that means the high-resolution inputs do not contain much extra information compared with low-resolution inputs. The interpolation operations would even cause additional errors on predictions. Thus, depth fusion method are not effective on low-resolution datasets, and the test results also show our performance on NYU and KITTI is not better than monocular backbones.
		
		\subsubsection{Running time}
		Detailed running time and computational statistics of monocular depth backbones and depth fusion methods are in Table~\ref{tab:ruuningtime}. 


	\end{appendices}
\end{document}